%% file: main.tex
\documentclass{article}

\PassOptionsToPackage{numbers}{natbib}
\usepackage[preprint]{neurips_2026}
\usepackage[utf8]{inputenc}
\usepackage[T1]{fontenc}
\usepackage{hyperref}
\usepackage{url}
\usepackage{booktabs}
\usepackage{amsfonts}
\usepackage{amsmath}
\usepackage{nicefrac}
\usepackage{microtype}
\usepackage[table]{xcolor}
\usepackage{diagbox}
\usepackage{graphicx}
\usepackage{multirow}
\usepackage{wrapfig}
\usepackage{float}
\definecolor{rowgray}{gray}{0.93}
\newcommand{\best}[1]{\textcolor{red}{#1}}

\title{BrainWorld: A Structural-Prior-Conditioned Generative Model for Whole-Brain 4D fMRI Dynamics}

\author{%
\textbf{Junfeng Xia\textsuperscript{1} \quad Wenhao Ye\textsuperscript{1,2} \quad Junxiang Zhang\textsuperscript{1} \quad Xuanye Pan\textsuperscript{1}}\\
\textbf{Mo Wang\textsuperscript{1,*} \quad Quanying Liu\textsuperscript{1,*}}\\
\textsuperscript{1}Department of Biomedical Engineering, Southern University of Science and Technology, China\\
\textsuperscript{2}School of Biomedical Engineering, Shenzhen University, China\\
\texttt{12250099@mail.sustech.edu.cn; liuqy@sustech.edu.cn}\\
\textsuperscript{*}Co-corresponding authors
}

\begin{document}

\maketitle

\begin{abstract}
Whole-brain 4D fMRI generation is valuable for modeling functional brain dynamics, yet existing fMRI foundation models mainly target representation learning and downstream prediction rather than conditional predictive generation. 
We introduce BrainWorld, a structural-prior-conditioned generative model for whole-brain 4D fMRI dynamics. 
BrainWorld uses sMRI as subject-level anatomical context to guide future fMRI generation, integrating structural information into the denoising process rather than treating it as a parallel modality. 
Evaluated on 22 datasets spanning diverse cohorts and brain states, BrainWorld generates stable 4D fMRI trajectories up to 400 frames, improves downstream performance through generated-example augmentation, and learns transferable multimodal representations that outperform baselines. 
Together, these results establish BrainWorld as a condition-aware generative framework for long-horizon brain dynamics modeling and multimodal representation learning. 
Code is available at this \href{https://github.com/REDMAO4869/BrainWorld}{link}.
\end{abstract}

\section{Introduction}
\begin{wrapfigure}[20]{r}{0.54\textwidth}
    \vspace{-1.95\baselineskip}
    \centering
    \includegraphics[width=\linewidth]{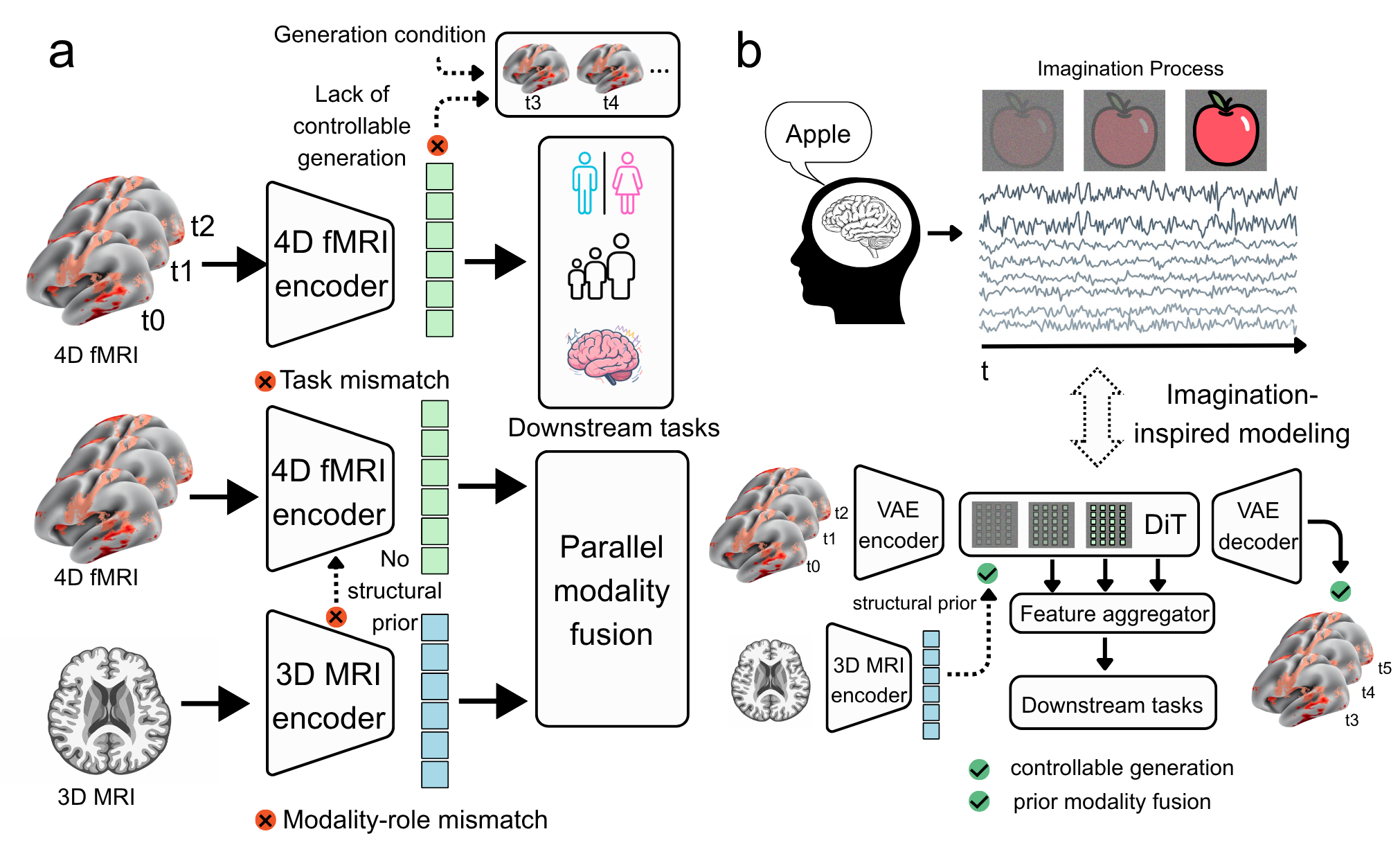}
    \caption{Motivation. Inspired by the process of imagination, BrainWorld addresses the task and modality-fusion mismatches by using sMRI as a conditioning input, guiding the generation of fMRI. As the brain refines vague concepts (e.g., an apple) into detailed features, BrainWorld leverages sMRI to transform the initial vague state into accurate predictions, naturally facilitating modality fusion for downstream tasks.}
    \label{fig:motivation}
\end{wrapfigure}

Imagination can be viewed as a generative process, in which coherent internal experiences are assembled from partial cues and prior knowledge~\cite{azizi2013computational}. World-model-style generative modeling offers a related perspective, focusing on how observations unfold over time under context and conditions~\cite{ha2018world}. This perspective is particularly relevant to functional magnetic resonance imaging (fMRI), because fMRI is not a static image but a temporally evolving observation of brain activity. In contrast, structural magnetic resonance imaging (sMRI) captures relatively stable anatomical organization, making it closer to static image modeling. Therefore, whole-brain 4D fMRI generation can be naturally formulated as condition-aware spatiotemporal generation, where future functional dynamics are modeled under both structural and contextual information.

While generative modeling has achieved notable success in medical imaging, particularly for the synthesis and completion of sMRI~\cite{pinaya2023generative}, fMRI generation remains far less explored, especially for condition-aware long-horizon generation of voxel-level whole-brain 4D signals. In practice, what is often scarce is not additional static structural images, but high-quality, long-horizon functional dynamics. This is especially true in rare diseases, small-sample task paradigms, and specialized stimulus-driven settings, where long-duration fMRI acquisition is costly and difficult. Consequently, condition-aware generation of whole-brain 4D fMRI is valuable not only for data synthesis, but also for learning informative dynamic representations and supporting condition-aware modeling of brain dynamics.

However, existing fMRI foundation models, although increasingly effective at capturing spatiotemporal patterns in large-scale brain data, are still primarily developed for representation learning and downstream transfer rather than for condition-aware long-horizon generation of voxel-level whole-brain 4D dynamics. In particular, most are centered on proxy objectives such as masked reconstruction, predictive representation learning, and contrastive learning, with the primary goal of learning informative representations for downstream tasks. Even recent generative formulations are mostly developed at the ROI level~\cite{xia2026brain}, rather than directly targeting condition-aware long-horizon generation of voxel-level whole-brain 4D signals. A further step is taken by structural--functional multimodal modeling, which introduces anatomical information into the learning process. Yet most existing approaches still treat the two modalities as parallel or complementary inputs~\cite{harmony,bi2024multimodal,khajehnejad2025brainsymphony}, rather than assigning sMRI an asymmetric role as subject-level anatomical context for functional dynamics generation. Together, these limitations point to the need for a modeling route that directly targets condition-aware long-horizon generation of whole-brain fMRI dynamics while naturally integrating structural MRI into functional modeling.

In this context, recent progress in video generation provides a promising modeling route for whole-brain 4D fMRI, as both tasks require long-horizon synthesis of high-dimensional spatiotemporal trajectories~\cite{brooks2024video,kong2024hunyuanvideo,wang2025wan}. Diffusion Transformers (DiTs)~\cite{DIT} are appealing for whole-brain 4D fMRI generation, as they combine diffusion-based modeling of complex high-dimensional distributions with Transformer-based long-range dependency modeling and flexible condition injection. Importantly, different denoising stages and intermediate DiT layers capture information at multiple levels, from coarse global dynamics to fine-grained local details. This allows the model to function not only as a high-quality generator, but also as a source of informative neural representations~\cite{lan2025diffusion,mukhopadhyay_text-free_2024,yang_diffusion_2023,wang_eegdm_2025,xia2026brain}. This generative framework also offers a more natural formulation for structural-functional integration. Since brain function evolves within subject-specific anatomy, sMRI can play an asymmetric role as anatomical context, rather than being treated as another parallel input modality. Compared with shallow concatenation or post-hoc feature pooling, injecting structural information into the denoising process allows anatomical context to modulate the formation of functional dynamics. Based on these observations, we propose BrainWorld, a structural-prior-conditioned generative model for whole-brain 4D fMRI dynamics. BrainWorld uses a variational autoencoder to compress voxel-level fMRI into a continuous latent space, where a conditional DiT models brain dynamics under sMRI conditioning, dynamic functional context, and optional stimulus conditions. This design supports condition-aware generation, dynamic representation learning, and generative structure-function fusion.

\noindent\textbf{Our main contributions are threefold:}
\begin{itemize}
    \setlength{\itemsep}{0.15em}
    \setlength{\parsep}{0pt}
    \setlength{\parskip}{0pt}
    \setlength{\topsep}{0.25em}
    \setlength{\partopsep}{0pt}

    \item We propose BrainWorld, a latent diffusion framework for long-horizon whole-brain 4D fMRI generation across diverse cohorts and brain states.

    \item We introduce a structural-prior conditioned denoising strategy that injects sMRI as subject-level anatomical context throughout the DiT backbone.

    \item We demonstrate BrainWorld's utility through three complementary evaluations: long-horizon predictive generation on four datasets, generated-example augmentation on two datasets, and representation transfer on downstream tasks.
\end{itemize}

\section{Related Work}

\subsection{fMRI Foundation Models}

Existing fMRI foundation models can be broadly organized by their proxy objectives. Reconstruction-based methods are relatively easy to scale and optimize, but remain largely focused on signal recovery~\cite{qu_uncovering_2024,wang2026omni,caro_brainlm_2023,wang_towards_2025}. JEPA-style methods replace direct reconstruction with predictive representation learning~\cite{dong2024brain,wang_slim-brain_2026,wang2026flexibrain}, emphasizing abstract latent prediction while requiring careful design to avoid collapse. Contrastive and LLM-alignment methods encourage transferable or semantically grounded latent representations, but are less directly aligned with explicit generative modeling of brain dynamics~\cite{yang_brainmass_2024,kim_SWIFT_2023,wei_fmri-lm_2025}. More recently, diffusion-based objectives have been explored as generative proxies for fMRI modeling, although current formulations remain mainly atlas-based, temporally restricted, and without structural conditioning~\cite{xia2026brain}. Taken together, these lines have substantially advanced fMRI representation learning and generative pretraining, but they are not directly designed for structural-condition-aware long-horizon generation of voxel-level whole-brain 4D fMRI.

\begin{figure*}[t]
    \centering
    \includegraphics[width=\linewidth]{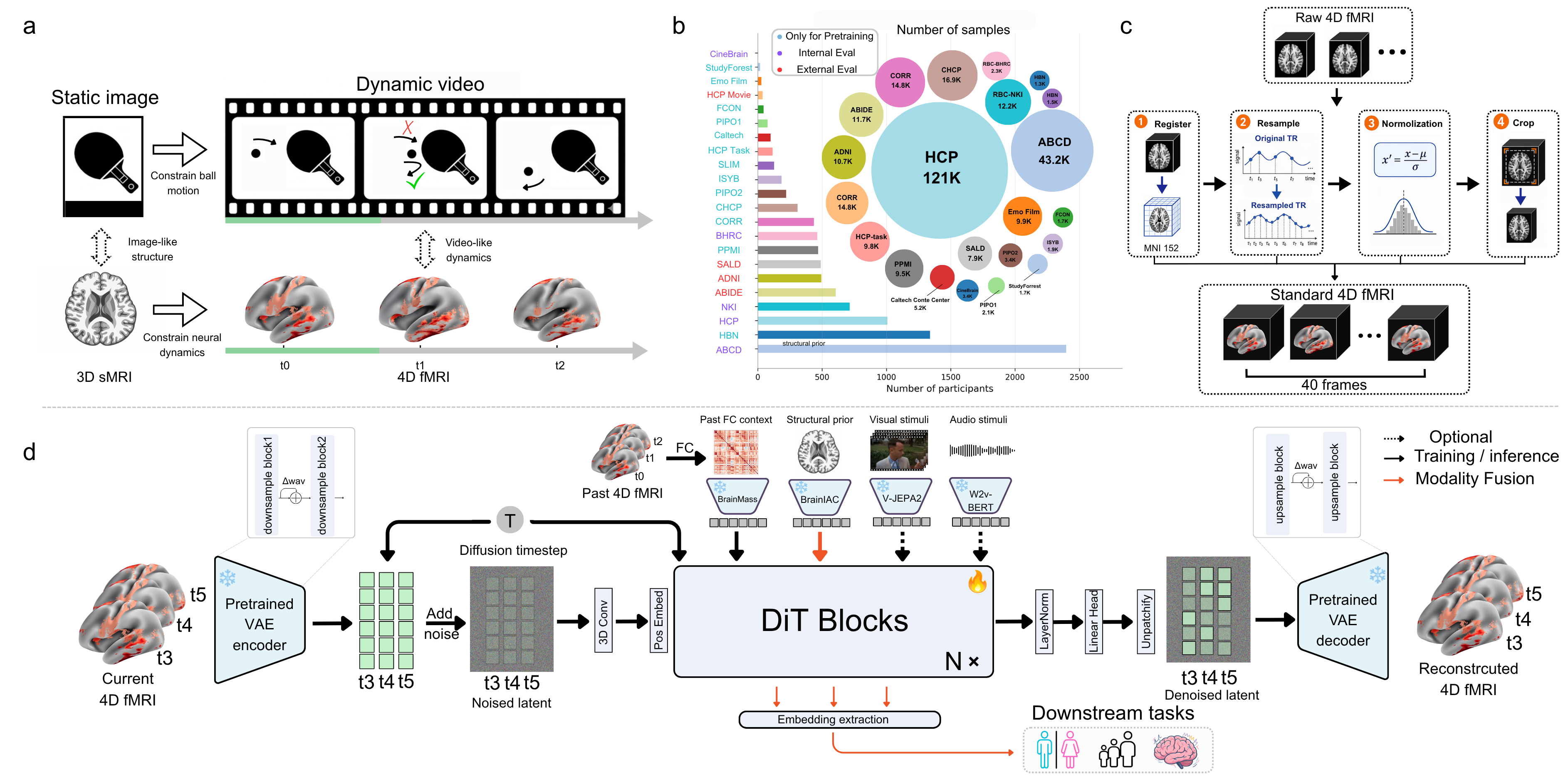}
    \caption{Overview.
(a) Motivation. sMRI resembles a static image that provides an anatomical scaffold, whereas 4D fMRI resembles a dynamic video of brain activity. Like a racket influencing a ball's trajectory, sMRI provides anatomical context for fMRI dynamics, motivating structural-condition-aware generation.
(b) Dataset. BrainWorld uses 22 fMRI datasets, grouped into pretraining-only, internal evaluation, and external evaluation cohorts.
(c) fMRI preprocessing. Raw 4D fMRI is standardized by MNI152 registration, TR resampling, signal normalization, and spatial cropping, yielding 40-frame inputs.
(d) Architecture. BrainWorld encodes voxel-level 4D fMRI into a pretrained VAE latent space and trains a DiT to denoise latent brain dynamics under past functional connectivity, structural MRI, and optional video/audio conditions encoded by frozen encoders. The model supports condition-aware 4D fMRI generation and downstream representation learning via intermediate DiT features.}
    \label{fig:framework}
\end{figure*}

\subsection{Medical Image Generation and fMRI Synthesis}

In medical image generation, sMRI and CT have established mature research paradigms~\cite{pinaya2023generative}. These models offer not only strong generative capability, but also useful intermediate representations for downstream tasks, highlighting the broader promise of generative pretraining~\cite{gao2025lung}. In contrast, fMRI generation remains less explored, especially for methods that directly model voxel-level whole-brain 4D spatiotemporal dynamics. Most prior work avoids the full complexity of 4D fMRI and instead focuses on simplified targets, such as ROI time series~\cite{hu2025synthesizing}, functional connectivity matrices~\cite{orlichenko2024demographic}, or short-horizon voxel-level prediction on limited-scale datasets~\cite{sun2025voxel}. Recently, Seo et al.~\cite{seo_scalable_2025} made an important step toward conditional whole-brain 4D fMRI synthesis. However, their formulation is primarily designed for task-specific conditional generation, leaving broader cross-state modeling of brain dynamics with subject-level structural conditioning less explored.

\subsection{Structural-Functional Joint Modeling}

Beyond fMRI generation itself, another relevant line concerns how structural information is integrated with functional dynamics. Structural-functional joint modeling has evolved from network-level fusion to fine-grained multimodal learning and, more recently, large-scale multimodal pretraining. Early studies fused structural and functional information through graph modeling or feature fusion to improve disease classification and diagnosis~\cite{zhang2021deep}. Later methods directly combined 3D sMRI and 4D fMRI, using co-attention, patch-level alignment, or latent contrastive objectives to mitigate cross-modal heterogeneity~\cite{bi2024multimodal,wei20264d}. Unlike earlier approaches trained end-to-end on relatively small datasets, BrainSymphony jointly models fMRI time series with diffusion-derived structural connectivity, while Brain Harmony compresses T1 morphology and fMRI dynamics into shared 1D tokens; both move toward multimodal foundation modeling through large-scale pretraining~\cite{khajehnejad2025brainsymphony,harmony}. Nevertheless, these approaches largely treat structure and function as parallel or complementary inputs. In contrast, BrainWorld uses sMRI as subject-level anatomical context to condition fMRI dynamics generation, motivating generative structure-function fusion for downstream tasks.

\section{Method}

\subsection{Overview}

Given a current whole-brain 4D fMRI window and aligned auxiliary conditions, BrainWorld generates future fMRI dynamics while extracting transferable representations from the generative process. To avoid costly voxel-space modeling, we use a two-stage latent framework. First, a 4D fMRI variational autoencoder (VAE) is pretrained on large-scale multi-cohort data to produce a compact and reconstructable latent space. Second, a conditional Diffusion Transformer (DiT) models future latent dynamics under functional, structural, and stimulus-related conditions. At inference, future latent windows are autoregressively generated from the current window and available conditions, then decoded back to voxel-level 4D fMRI by the pretrained VAE decoder. Intermediate DiT states are further used as representations for downstream tasks, supporting both condition-aware generation and foundation-model-style representation learning.

\subsection{Stage I: 4D fMRI VAE Pretraining}

We first pretrain a frame-wise spatial VAE for 4D fMRI windows to provide a compact and reconstructable latent space for conditional diffusion modeling. To reduce background-dominated computation, each standardized \(96^3\) fMRI volume is cropped to remove non-brain background regions, yielding an input window \(x_i\in\mathbb{R}^{40\times80\times96\times80}\). The VAE preserves the temporal axis and applies a shared 3D encoder frame by frame. With \(C=16\) latent channels, the resulting latent sequence is
\[
u_i\in\mathbb{R}^{40\times16\times10\times12\times10}.
\]

The encoder uses three spatial downsampling stages to reduce each frame from \(80\times96\times80\) to \(10\times12\times10\), while the decoder symmetrically uses three upsampling stages to reconstruct the original cropped resolution. Following the design of WF-VAE~\cite{WF_VAE}, we incorporate lightweight 3D Haar wavelet fusion blocks into intermediate features, where fixed Haar transforms capture low- and high-frequency spatial cues and residual fusion integrates them back into the feature stream. The VAE is trained with
\begin{equation}
\mathcal{L}_{\mathrm{VAE}}
=
\mathcal{L}_{\mathrm{rec}}
+
\lambda_{\mathrm{bg}}\mathcal{L}_{\mathrm{bg}}
+
\lambda_{\mathrm{wav}}\mathcal{L}_{\mathrm{wav}}
+
\beta\mathcal{L}_{\mathrm{KL}}.
\end{equation}
The four terms are defined as
\begin{equation}
\begin{aligned}
\mathcal{L}_{\mathrm{rec}} &=
\frac{\sum_v w_v |x_v-\hat{x}_v|}{\sum_v w_v}, \quad
\mathcal{L}_{\mathrm{bg}} =
\frac{1}{|\Omega_{\mathrm{bg}}|}\sum_{v\in\Omega_{\mathrm{bg}}}|\hat{x}_v|, \\
\mathcal{L}_{\mathrm{wav}} &=
\|H_{\mathrm{haar}}(x)-H_{\mathrm{haar}}(\hat{x})\|_1, \quad
\mathcal{L}_{\mathrm{KL}} =
\frac{1}{2}\mathbb{E}\left[\mu^2+\sigma^2-\log\sigma^2-1\right].
\end{aligned}
\end{equation}
Here, \(w_v\) upweights foreground voxels, \(\Omega_{\mathrm{bg}}\) denotes the background region, and \(H_{\mathrm{haar}}\) is applied frame-wise to enforce 3D wavelet-domain consistency.
\begin{figure*}[t]
    \centering
    \includegraphics[width=\linewidth]{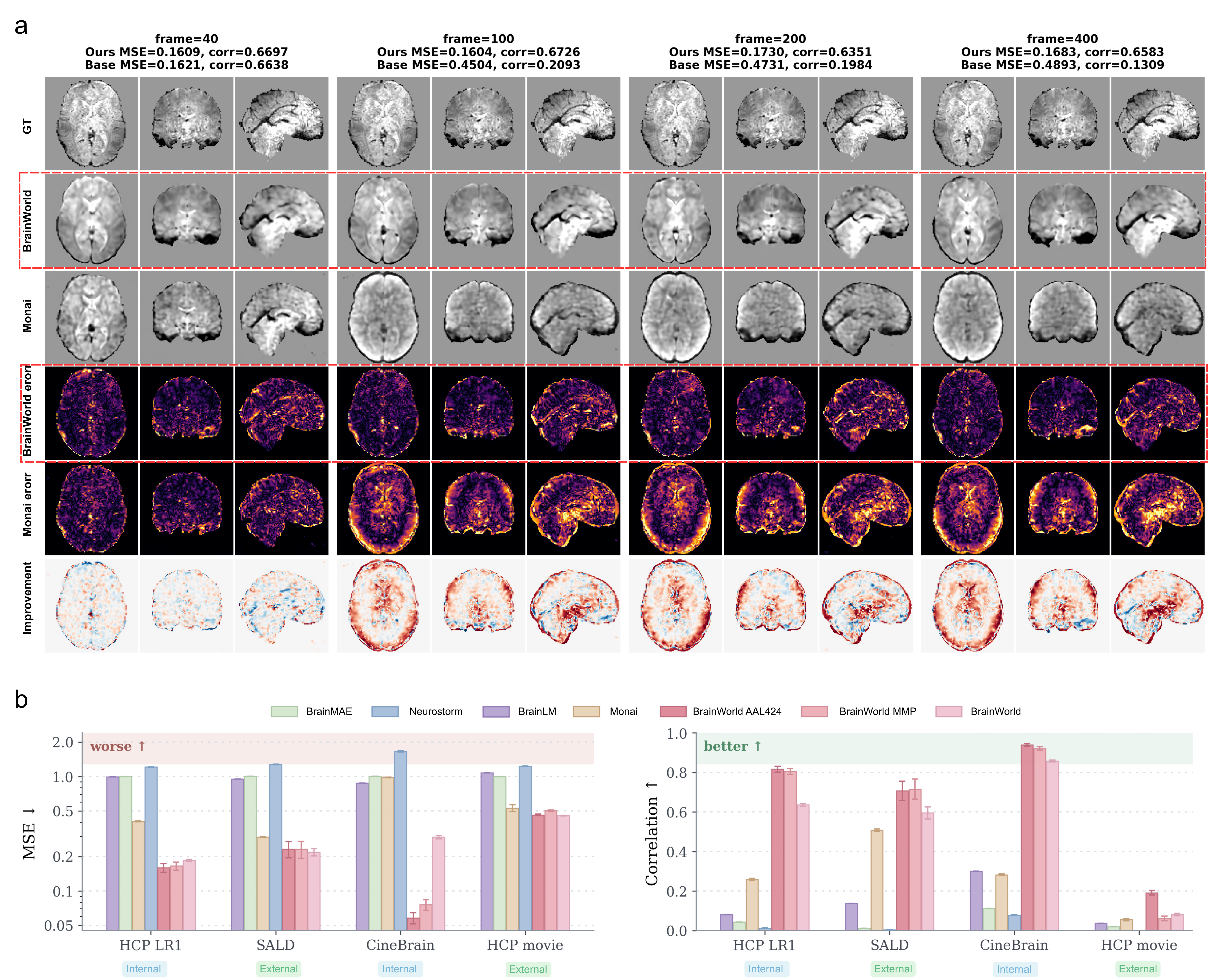}
    \caption{
    Long-horizon 4D fMRI generation and cross-dataset evaluation.
    (a) Representative HCP subject visualization at 40, 100, 200, and 400 predicted frames, comparing GT, BrainWorld, MONAI, error maps, and voxel-wise improvements.
    (b) Quantitative results across internal/external datasets and resting-state/visual-stimulus fMRI.
    BrainWorld-MMP and BrainWorld-AAL424 denote BrainWorld predictions parcellated using the MMP~\cite{MMP} and AAL424~\cite{AAL424} atlases, respectively.
    Additional visualizations and generation metrics across different prediction horizons are provided in Supplementary Sections~\ref{supp:Generation_Visualizations} and~\ref{supp:Generation_Statistics}.
    }
    \label{fig:generation}
\end{figure*}

\subsection{Stage II: Conditional Latent Diffusion Modeling}

After VAE pretraining, BrainWorld learns a conditional latent diffusion model
\(p_\theta(u_{i+1}\mid u_i,c_i)\), where \(u_i\) and \(u_{i+1}\) are the current and next fMRI-window latents. 
The condition \(c_i\) includes past FC, sMRI, and optional video/audio embeddings as dynamic, anatomical, and stimulus contexts. 
Here, ``structural prior'' denotes sMRI-based conditioning rather than an explicit Bayesian prior. 
All conditions are extracted by frozen encoders and projected into the DiT conditioning space, with details in Supplementary Section~\ref{supp:condition-preparation}.

We formulate next-window prediction as denoising the future latent. 
The clean target is \(u_0:=u_{i+1}\), and the forward diffusion process is
\begin{equation}
u_t=\sqrt{\bar{\alpha}_t}u_0+\sqrt{1-\bar{\alpha}_t}\epsilon,
\quad \epsilon\sim\mathcal{N}(0,I),
\end{equation}
where \(t\in\{1,\ldots,T_{\mathrm{diff}}\}\). 
We use \(T_{\mathrm{diff}}=1000\) diffusion steps with a cosine noise schedule. 
The noisy latent \(u_t\) is spatially patchified into spatiotemporal tokens while preserving the temporal axis, enabling the DiT to model both spatial dependencies and temporal evolution.

Conditions are injected through complementary token-level and global pathways. 
At the token level, condition embeddings are incorporated by cross-attention to provide content-dependent modulation of latent tokens. 
At the global level, condition embeddings are fused with the diffusion timestep embedding and used to modulate each DiT block through adaptive normalization and residual gating. 
This allows sMRI, functional context, and stimulus information to modulate denoising throughout the network, rather than being fused only at the input or output.

We train the DiT with \(v\)-prediction:
\begin{equation}
v=\sqrt{\bar{\alpha}_t}\epsilon-\sqrt{1-\bar{\alpha}_t}u_0,
\quad
\mathcal{L}_{v}
=
\left\|
\hat{v}_{\theta}(u_t,t,u_i,c_i)-v
\right\|_2^2 .
\end{equation}
To stabilize clean-latent recovery, we reconstruct \(\hat{u}_0\) from the predicted \(v\) and add an auxiliary loss:
\begin{equation}
\mathcal{L}_{\mathrm{StageII}}
=
\mathcal{L}_{v}
+
\lambda_{x_0}\mathrm{SmoothL1}(\hat{u}_0,u_0),
\quad
\lambda_{x_0}=0.05 .
\end{equation}

At inference, the predicted latent is decoded by the frozen VAE decoder and autoregressively rolled out for long-horizon generation; intermediate DiT states are used for downstream representation extraction.
\subsection{Autoregressive Latent Rollout and Generation}

To generate trajectories longer than one 40-frame block, we autoregressively unroll the one-step latent predictor learned in Stage II. Given the current latent window \(u_i\) and condition set \(c_i\), BrainWorld samples the next latent block
\[
\hat{u}_{i+1}\sim p_\theta(u_{i+1}\mid u_i,c_i),
\]
and reconstructs it into voxel-level 4D fMRI using the frozen VAE decoder:
\[
\hat{x}_{i+1}=D_\psi(\hat{u}_{i+1}).
\]

During rollout, different conditions follow different update rules. Specifically, each generated voxel-level block is parcellated into Schaefer-100 ROI time series~\cite{Schaefer}, from which an ROI-wise Pearson FC matrix is computed, vectorized, projected into an 800-D embedding, and used as the functional context for the next rollout step. sMRI serves as a subject-level anatomical prior and remains fixed across the rollout. Optional video/audio conditions are externally aligned to the corresponding future time intervals when available. For a target horizon of \(H\) frames and block length \(B=40\), we perform
\[
K=\left\lceil \frac{H}{B} \right\rceil
\]
rollout steps and concatenate the generated blocks in temporal order. This enables efficient long-horizon generation by performing denoising in latent space while updating functional context from generated 4D fMRI.

\subsection{Modality-Fused Representation Extraction for Downstream Tasks}

Beyond generation, we use the trained DiT as a diffusion-based representation encoder. 
For each downstream sample, the observed fMRI window \(x_i\) is encoded by the pretrained VAE into a latent window \(u_i\), which is then fed into the DiT to extract intermediate features paired with the label of the same window. 
No future window, generated sample, or FC embedding is used for downstream supervision or feature extraction.

For each selected diffusion timestep \(t\), the current latent window \(u_i\) is perturbed using the Stage-II noise schedule and fed into the DiT. Hidden states from the selected timesteps and layers are then collected as multi-scale representations.
Let \(H_{t,l}\in\mathbb{R}^{N\times D}\) denote the hidden tokens at timestep \(t\) and layer \(l\). 
We mean-pool over tokens and concatenate features across selected timesteps and layers:
\[
e_{t,l}=\frac{1}{N}\sum_{n=1}^{N}H_{t,l,n}, 
\quad
e=\mathrm{Concat}_{t\in\mathcal{T}_{\mathrm{feat}},\,l\in\mathcal{L}_{\mathrm{feat}}}(e_{t,l}).
\]
We use a fixed set of diffusion timesteps and late DiT layers for feature extraction, with detailed settings provided in Supplementary Table~\ref{tab:ddit-configs}.

To test how sMRI should be integrated, we compare three variants: BrainWorld$_{\mathrm{F}}$, which extracts features from fMRI latents only; BrainWorld$_{\mathrm{F+S\text{-}concat}}$, which concatenates fMRI features and independently extracted sMRI embeddings before the downstream head; and BrainWorld$_{\mathrm{F+S\text{-}prior}}$, which injects sMRI into the DiT through the structural-prior conditioning pathway. 
These variants distinguish post-hoc representation fusion from structural-prior fusion within the DiT.

\section{Experiments}

\subsection{Experimental Setup}

\paragraph{Datasets and Preprocessing.}
As shown in Figure~\ref{fig:framework}, we use 22 public datasets spanning resting-state, task-driven, naturalistic movie-based, and clinical cohorts~\cite{gao_cinebrain_2025,StudyForrest,Emo_Film,HCP,caltech,SLIM,ISYB,PIPO1_PIPO2,CHCP,CORR,BHRC,PPMI,SALD,ADNI,ABIDE,NKI,ABCD,HBN,FCON}. The datasets are divided into pretraining-only, internal evaluation, and external evaluation splits to assess generalization under distribution shifts. All fMRI scans are spatially aligned to a common template, resampled to 2 mm isotropic resolution, and temporally interpolated to a unified TR of 0.72 s. Voxel intensities are normalized using background-aware \(Z\)-scoring. Each volume is then cropped or padded to a fixed spatial size of \(96\times96\times96\), and each time series is segmented into 40-frame windows. For MRI, images are skull-stripped, bias-field corrected, and resampled to 1 mm MNI152 space, serving as anatomical conditioning inputs. Details of the datasets and preprocessing pipelines are provided in Supplementary Sections~\ref{supp:datasets} and~\ref{supp:processing}, respectively.

\paragraph{Baselines and Evaluation Protocols.}

We evaluate BrainWorld on two task families: generative forecasting and downstream transfer. 
For generative forecasting, we compare against representative fMRI foundation models, including BrainMAE (an ROI-level predictive baseline implemented for controlled comparison; see Supplementary Section~\ref{supp:BrainMAE}), BrainLM~\cite{caro_brainlm_2023}, and NeuroSTORM~\cite{wang_towards_2025}. 
We also include a MONAI-based latent diffusion baseline~\cite{pinaya2023generative} adapted for 4D fMRI prediction (see Supplementary Section~\ref{supp:monai_intro}).
For fair quantitative comparison, voxel-level predictions from BrainWorld and voxel baselines are evaluated in voxel space or parcellated to the corresponding ROI space when compared with ROI-level baselines.
For downstream transfer, we compare functional-only (F), structural-only (S), and multimodal (F+S) baselines under two protocols: full fine-tuning for end-to-end adaptability and linear probing for frozen-backbone representation transfer. 
Detailed baseline descriptions, computational resources, and evaluation protocols are provided in Supplementary Sections~\ref{supp:baseline},~\ref{supp:compute}, and~\ref{supp:evaluation-details}, respectively.

\subsection{BrainWorld Enables Long-Horizon Predictive Generation of 4D fMRI}

We evaluate BrainWorld as a long-horizon 4D fMRI predictive generator across resting-state and stimulus-driven conditions. Figure~\ref{fig:generation}(a) visualizes a representative HCP subject at prediction horizons of 40, 100, 200, and 400 frames, comparing ground truth, BrainWorld, the MONAI baseline, error maps, and voxel-wise improvement maps. BrainWorld preserves clearer whole-brain structure and lower voxel-level error across horizons, with the advantage remaining evident at 200 and 400 frames. In contrast, MONAI progressively loses anatomical detail and accumulates larger errors as the prediction horizon increases. 

Figure~\ref{fig:generation}(b) reports 400-frame quantitative results across four settings: internal resting-state HCP, external resting-state SALD, internal visual-stimulus CineBrain, and external visual-stimulus HCP Movie. BrainWorld consistently obtains lower MSE and higher Pearson correlation than competing baselines across these settings, supporting robust long-horizon predictive generation across both brain states and dataset sources. Its strong performance on SALD and HCP Movie further suggests generalization to unseen cohorts and acquisition paradigms.
\begin{figure*}[t]
    \centering
    \includegraphics[width=\linewidth]{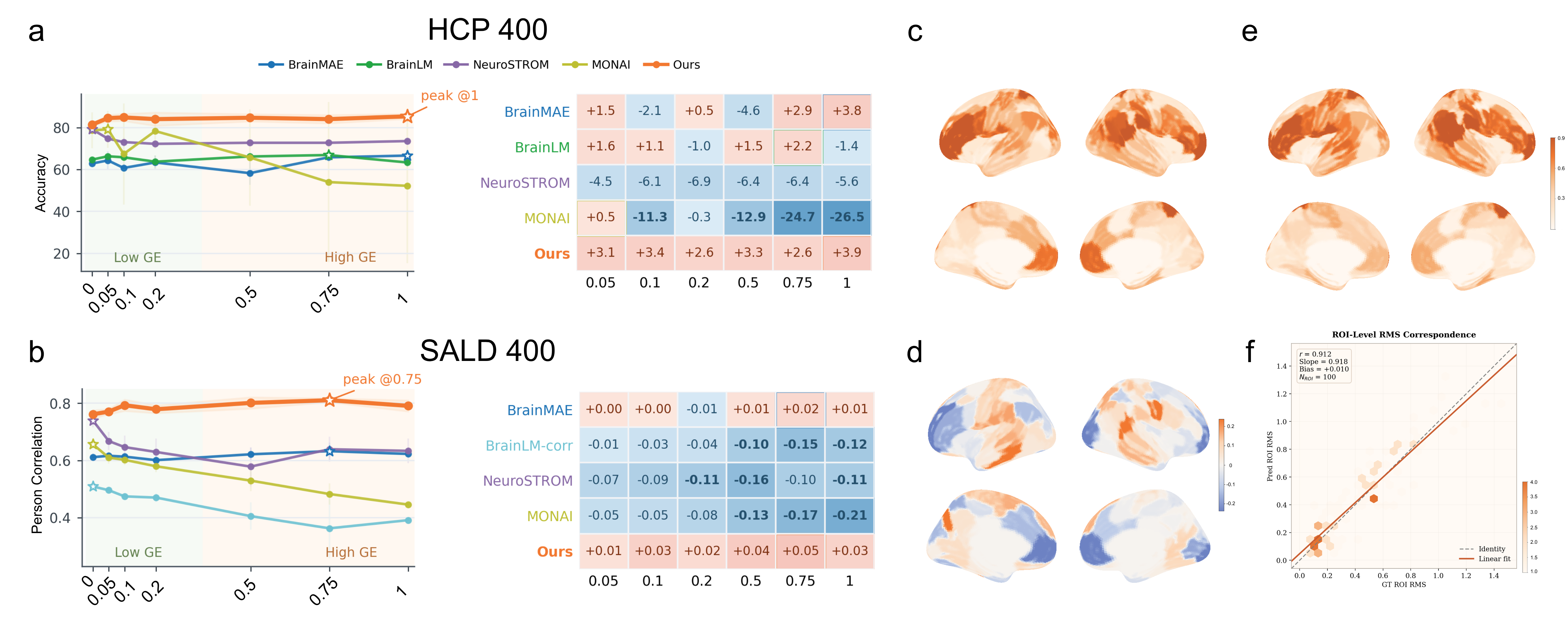}
    \caption{
    Generated-example (GE) scaling improves downstream performance while preserving ROI-level response structure at a 400-frame horizon. 
    (a) HCP-400 gender classification results across GE ratios, where higher ratios add more generated examples. 
    The line plot shows absolute accuracy, and the heatmap shows gains over each model's \(\mathrm{GE}=0\) baseline in percentage points. 
    (b) SALD-400 age regression results in the same format, with gains reported in Pearson correlation units. 
    (c,e) Cohort-level ground-truth and predicted ROI RMS surface maps on CineBrain. 
    (d) Prediction-minus-ground-truth ROI RMS difference maps. 
    (f) ROI-wise correspondence between predicted and ground-truth RMS values across Schaefer-100 parcels~\cite{Schaefer}.
    }
    \label{fig:augmentation}
\end{figure*}

\subsection{Generated fMRI Supports Downstream Augmentation}

To evaluate whether generated data improves downstream performance, we conduct generated-example (GE) augmentation with varying GE ratios while keeping the real training set fixed. Generated examples inherit labels and sMRI conditions from their conditioning real samples, and are used as label-preserving augmentations rather than independent biological measurements. All GE ratios use the same optimization steps and validation-based early stopping, and are evaluated on held-out real test sets. As shown in Fig.~\ref{fig:augmentation}(a,b), HCP-400 refers to gender classification on HCP using 400-frame samples, while SALD-400 refers to age regression on SALD. \(\mathrm{GE}=0\) uses only real data, while \(\mathrm{GE}=1.0\) adds an equal number of generated and real samples. BrainWorld achieves the best performance across all GE ratios on both tasks. On HCP-400, accuracy improves from \(81.45\%\) to \(85.39\%\) at \(\mathrm{GE}=1.0\) (\(+3.94\) points), and on SALD-400, Pearson correlation increases from \(0.7613\) to \(0.8104\) at \(\mathrm{GE}=0.75\) (\(+0.0491\)). In contrast, the strongest baseline improves by only \(+0.0210\), with most baselines degrading as GE increases.

To assess spatial fidelity, we evaluate Schaefer-100 ROI temporal RMS on 38 unseen CineBrain runs, as shown in Fig.~\ref{fig:augmentation}(c--f). The predicted RMS closely matches ground truth (\(r=0.912\), cosine similarity \(=0.973\), slope \(=0.918\), bias \(=+0.010\)), with localized errors. This suggests that BrainWorld preserves ROI-wise functional response patterns in generated samples, which may explain their utility for downstream augmentation.

\subsection{Structural-Prior Conditioning Improves Multimodal Representation Transfer}

We explore whether sMRI is more effective when injected as a structural conditioning signal than when used as a parallel feature for multimodal fusion. To this end, we compare BrainWorld$_{\mathrm{F+S\text{-}prior}}$ with functional-only (F), structural-only (S), and multimodal fusion baselines, including Brain Harmony as a representative representation-level fusion method. To further isolate the fusion mechanism from backbone differences, we introduce a same-backbone concat baseline, BrainWorld$_{\mathrm{F+S\text{-}concat}}$, which independently extracts fMRI and sMRI features and concatenates them before the downstream head. BrainWorld$_{\mathrm{F}}$ denotes the corresponding fMRI-only variant without sMRI conditioning. Two transfer protocols are used: full fine-tuning, which evaluates end-to-end downstream adaptability, and linear probing with a frozen backbone, which evaluates representation transferability. For age regression, target ages are z-score normalized within the training set. Given the limited number of runs ($n=3$), Cohen's $d$~\cite{cohen} is used to quantify the effect size of performance differences.

Table~\ref{tab:ft} shows that BrainWorld$_{\mathrm{F+S\text{-}prior}}$ achieves the best full fine-tuning performance across five datasets, demonstrating strong adaptability to diverse demographic tasks. This advantage also holds under the frozen-backbone protocol in Table~\ref{tab:lp}, indicating that denoising-time structural conditioning improves not only task-specific adaptation but also the transferability of learned representations. Notably, BrainWorld$_{\mathrm{F+S\text{-}concat}}$ does not consistently improve over BrainWorld$_{\mathrm{F}}$, suggesting that sMRI is not automatically beneficial when used as a post-hoc feature. In contrast, BrainWorld$_{\mathrm{F+S\text{-}prior}}$ achieves the strongest overall performance, supporting the benefit of injecting sMRI into the DiT denoising pathway rather than simply concatenating it as a post-hoc representation.
\begin{table*}[t]
\centering 
\small
\setlength{\tabcolsep}{2pt} 
\renewcommand{\arraystretch}{1.15}
\caption{
Full fine-tuning performance on demographic and disease diagnosis tasks. 
We report ACC/F1 for classification and MSE/Pearson correlation for regression.
F, S, and F+S denote functional-only, structural-only, and multimodal models, respectively. 
\best{Red} indicates the best performance, and \underline{underlining} indicates the second-best performance.
\(^{*}\) denotes a large favorable effect size with Cohen's \(d \geq 0.8\) compared with the best-performing non-BrainWorld baseline for the same task and metric.
}
\label{tab:ft}
\resizebox{\textwidth}{!}{
\begin{tabular}{l c cc cc cc cc cc}
\toprule
\multirow{3}{*}{\diagbox[width=7em]{\textbf{Model}}{\textbf{Dataset}}} & \multirow{3}{*}{\textbf{Modality}} &
\multicolumn{2}{c}{\textbf{ABIDE}} & \multicolumn{2}{c}{\textbf{NKI}} & \multicolumn{2}{c}{\textbf{SALD}} & \multicolumn{2}{c}{\textbf{ABCD}} & \multicolumn{2}{c}{\textbf{BHRC}} \\
 & & \multicolumn{2}{c}{\textit{Age Regression}} & \multicolumn{2}{c}{\textit{Age Regression}} & \multicolumn{2}{c}{\textit{Age Regression}} & \multicolumn{2}{c}{\textit{Gender Classif.}} & \multicolumn{2}{c}{\textit{Gender Classif.}} \\
\cmidrule(lr){3-4} \cmidrule(lr){5-6}\cmidrule(lr){7-8}\cmidrule(lr){9-10}\cmidrule(lr){11-12}
 & & \textbf{MSE $\downarrow$ } & \textbf{R $\uparrow$ } & \textbf{MSE $\downarrow$ } & \textbf{R $\uparrow$ } & \textbf{MSE $\downarrow$ } & \textbf{R $\uparrow$ } & \textbf{ACC  $\uparrow$ } & \textbf{F1 $\uparrow$ } & \textbf{ACC  $\uparrow$ } & \textbf{F1 $\uparrow$ } \\
\midrule
BrainLM~\cite{caro_brainlm_2023} & F & 0.907$\pm$0.007 & 0.239$\pm$0.023 & 0.503$\pm$0.015 & 0.678$\pm$0.014 & 0.677$\pm$0.055 & 0.634$\pm$0.061 & 59.24$\pm$1.03 & 58.74$\pm$0.70 & 57.09$\pm$1.33 & 54.69$\pm$1.03 \\
Brain-JEPA~\cite{dong2024brain} & F & 0.982$\pm$0.066 & 0.166$\pm$0.017 & 1.030$\pm$0.080 & 0.330$\pm$0.030 & 1.130$\pm$0.108 & 0.299$\pm$0.060 & - & - & 58.16$\pm$6.10 & 56.16$\pm$6.05 \\
BrainMass~\cite{yang_brainmass_2024} & F & 0.695$\pm$0.038 & 0.500$\pm$0.044 & 0.600$\pm$0.062 & 0.620$\pm$0.036 & 0.704$\pm$0.080 & 0.626$\pm$0.062 & 57.92$\pm$1.19 & 57.86$\pm$1.18 & 62.76$\pm$1.50 & 61.05$\pm$1.22 \\
Brain-DiT~\cite{xia2026brain} & F & 0.750$\pm$0.010 & 0.610$\pm$0.010 & 0.380$\pm$0.010 & 0.810$\pm$0.010 & 0.330$\pm$0.010 & 0.790$\pm$0.010 & 61.67$\pm$1.15 & 60.01$\pm$3.40 & 56.60$\pm$2.35 & 48.86$\pm$5.85 \\
SWIFT~\cite{kim_SWIFT_2023} & F & 0.524$\pm$0.010 & 0.704$\pm$0.011 & 0.127$\pm$0.004 & 0.946$\pm$0.002 & 0.164$\pm$0.080 & \underline{0.926$\pm$0.003} & 66.67$\pm$3.51 & 68.69$\pm$0.44 & 70.21$\pm$0.02 & 69.56$\pm$0.02 \\
SlimBrain~\cite{wang_slim-brain_2026} & F & 0.493$\pm$0.012 & 0.285$\pm$0.015 & 0.667$\pm$0.083 & 0.613$\pm$0.078 & 0.705$\pm$0.054 & 0.590$\pm$0.023 & 67.92$\pm$1.10 & 67.18$\pm$1.76 & 71.74$\pm$2.17 & 69.35$\pm$4.25 \\
NeuroSTORM~\cite{wang_towards_2025} & F & 0.537$\pm$0.016 & 0.648$\pm$0.013 & 0.152$\pm$0.033 & 0.835$\pm$0.035 & 0.194$\pm$0.019 & 0.892$\pm$0.010 & 76.51$\pm$1.88 & 76.24$\pm$1.70 & 73.02$\pm$2.19 & 72.49$\pm$1.96 \\
Omni-fMRI~\cite{wang2026omni} & F & 0.427$\pm$0.006 & 0.734$\pm$0.002 & \underline{0.088$\pm$0.004} & \underline{0.959$\pm$0.003} & 0.182$\pm$0.019 & 0.909$\pm$0.005 & 77.01$\pm$0.55 & 76.91$\pm$0.54 & 75.86$\pm$1.35 & 75.96$\pm$1.31 \\
Brain Harmony-S~\cite{harmony} & S & 0.318$\pm$0.016 & 0.873$\pm$0.008 & 0.266$\pm$0.012 & 0.868$\pm$0.005 & 0.553$\pm$0.054 & 0.632$\pm$0.038 & 74.33$\pm$0.94 & 73.96$\pm$0.75 & 65.96$\pm$3.13 & 65.76$\pm$3.22 \\
Brain Harmony~\cite{harmony} & F + S & 0.373$\pm$0.028 & 0.753$\pm$0.023 & 0.171$\pm$0.016 & 0.822$\pm$0.018 & 0.188$\pm$0.017 & 0.667$\pm$0.013 & 75.17$\pm$2.08 & 75.16$\pm$2.09 & 69.50$\pm$1.23 & 68.74$\pm$1.17 \\
\midrule
BrainWorld$_{\mathrm{F}}$ & F & \underline{0.127$\pm$0.058} & \underline{0.915$\pm$0.026} & 0.163$\pm$0.036 & 0.949$\pm$0.007 & \underline{0.143$\pm$0.015} & 0.912$\pm$0.009 & \underline{79.43$\pm$0.45} & \underline{79.42$\pm$0.46} & \underline{78.37$\pm$2.92} & \underline{78.09$\pm$3.17} \\
BrainWorld$_{\mathrm{F+S\text{-}prior}}$ & F + S & \best{\textbf{0.062$\pm$0.011*}} & \best{\textbf{0.948$\pm$0.008*}} & \best{\textbf{0.083$\pm$0.001*}} & \best{\textbf{0.962$\pm$0.001*}} & \best{\textbf{0.107$\pm$0.007*}} & \best{\textbf{0.935$\pm$0.005*}} & \best{\textbf{80.75$\pm$1.71*}} & \best{\textbf{80.75$\pm$1.71*}} & \best{\textbf{79.94$\pm$2.07*}} & \best{\textbf{79.84$\pm$2.03*}} \\
\bottomrule
\end{tabular}
}
\end{table*}

\begin{table*}[t]
\centering 
\small
\setlength{\tabcolsep}{2pt} 
\renewcommand{\arraystretch}{1.15}
\caption{
Performance on demographic and disease diagnosis tasks using a frozen backbone. 
We report ACC/F1 for classification and MSE/Pearson correlation for regression.
F, S, and F+S denote functional-only, structural-only, and multimodal models, respectively. 
\best{Red} indicates the best performance and \underline{underline} indicates the second-best performance.
\(^{*}\) denotes a large favorable effect size with Cohen's \(d \geq 0.8\) compared with the best-performing non-BrainWorld baseline for the same task and metric.
}
\label{tab:lp}
\resizebox{\textwidth}{!}{
\begin{tabular}{l c cc cc cc cc}
\toprule 
\multirow{3}{*}{\diagbox[width=7em]{\textbf{Model}}{\textbf{Dataset}}} & \multirow{3}{*}{\textbf{Modality}} &
\multicolumn{2}{c}{\textbf{NKI}} & \multicolumn{2}{c}{\textbf{ABCD}} & \multicolumn{2}{c}{\textbf{SALD}} & \multicolumn{2}{c}{\textbf{ADNI (MCI)}} \\
 & & \multicolumn{2}{c}{\textit{Education Classif.}} & \multicolumn{2}{c}{\textit{Gender Classif.}} & \multicolumn{2}{c}{\textit{Age Regression}} & \multicolumn{2}{c}{\textit{Diagnosis}} \\
\cmidrule(lr){3-4} \cmidrule(lr){5-6}\cmidrule(lr){7-8}\cmidrule(lr){9-10}
 & & \textbf{ACC  $\uparrow$ } & \textbf{F1 $\uparrow$ } & \textbf{ACC  $\uparrow$ } & \textbf{F1 $\uparrow$ } & \textbf{MSE $\downarrow$ } & \textbf{R $\uparrow$ } & \textbf{ACC  $\uparrow$ } & \textbf{F1 $\uparrow$ } \\
\midrule
Brain-DiT~\cite{xia2026brain} & F & 62.80$\pm$0.63 & 60.26$\pm$1.22 & 55.47$\pm$0.03 & 52.37$\pm$0.39 & 0.666$\pm$0.005 & 0.495$\pm$0.007 & 47.03$\pm$0.49 & 46.94$\pm$0.54 \\
SWIFT~\cite{kim_SWIFT_2023} & F & 61.55$\pm$2.52 & 58.23$\pm$4.20 & 59.31$\pm$2.09 & 58.16$\pm$2.84 & 0.933$\pm$0.042 & 0.460$\pm$0.131 & 69.78$\pm$4.63 & 69.31$\pm$5.23 \\
SlimBrain~\cite{wang_slim-brain_2026} & F & 59.94$\pm$1.43 & 59.10$\pm$1.79 & 67.92$\pm$1.10 & 67.18$\pm$1.76 & 0.680$\pm$0.100 & 0.590$\pm$0.093 & 59.14$\pm$7.00 & 57.98$\pm$6.85 \\
NeuroSTORM~\cite{wang_towards_2025} & F & 66.78$\pm$0.51 & 57.37$\pm$0.86 & 63.06$\pm$6.49 & 60.76$\pm$10.16 & 0.386$\pm$0.035 & 0.774$\pm$0.004 & 57.25$\pm$3.18 & 56.76$\pm$3.21 \\
Omni-fMRI~\cite{wang2026omni} & F & 68.13$\pm$0.60 & 63.65$\pm$2.04 & \underline{73.96$\pm$1.80} & \underline{73.74$\pm$2.18} & 0.354$\pm$0.048 & \underline{0.814$\pm$0.024} & 63.40$\pm$2.55 & 56.67$\pm$9.60 \\
BrainIAC~\cite{BrainIAC} & S & 63.93$\pm$2.33 & 62.97$\pm$2.36 & 65.76$\pm$1.61 & 65.69$\pm$1.57 & 0.416$\pm$0.015 & 0.750$\pm$0.003 & \underline{73.73$\pm$0.69} & \underline{73.56$\pm$0.66} \\
Brain Harmony-S~\cite{harmony} & S & 82.10$\pm$5.72 & 82.09$\pm$5.19 & 69.87$\pm$1.07 & 67.09$\pm$0.80 & \underline{0.343$\pm$0.025} & 0.803$\pm$0.018 & 52.27$\pm$1.86 & 52.15$\pm$1.86 \\
Brain Harmony~\cite{harmony} & F + S & 81.48$\pm$0.01 & 80.66$\pm$0.85 & 70.13$\pm$3.44 & 69.42$\pm$3.21 & 0.386$\pm$0.034 & 0.802$\pm$0.025 & 58.33$\pm$3.86 & 58.28$\pm$4.01 \\
\midrule
BrainWorld$_{\mathrm{F}}$ & F & \underline{86.76$\pm$0.20} & \underline{86.27$\pm$0.17} & 69.81$\pm$0.07 & 69.80$\pm$0.08 & 0.380$\pm$0.020 & 0.792$\pm$0.008 & 67.82$\pm$0.59 & 67.77$\pm$0.53 \\
BrainWorld$_{\mathrm{F+S\text{-}concat}}$ & F + S & 81.62$\pm$1.96 & 81.53$\pm$1.84 & 67.31$\pm$0.62 & 67.25$\pm$0.64 & 0.442$\pm$0.040 & 0.754$\pm$0.007 & 67.36$\pm$2.99 & 67.31$\pm$3.13 \\
BrainWorld$_{\mathrm{F+S\text{-}prior}}$ & F + S & \best{\textbf{87.67$\pm$0.68*}} & \best{\textbf{87.21$\pm$0.80*}} & \best{\textbf{74.09$\pm$0.24}} & \best{\textbf{74.10$\pm$0.24}} & \best{\textbf{0.269$\pm$0.007*}} & \best{\textbf{0.830$\pm$0.005*}} & \best{\textbf{76.27$\pm$6.14}} & \best{\textbf{76.26$\pm$6.12}} \\
\bottomrule
\end{tabular}
}
\end{table*}

\section{Discussion}

Our results support formulating whole-brain 4D fMRI generation as structural-condition-aware brain dynamics modeling. BrainWorld achieves stable long-horizon predictive generation across datasets and brain states, and the generated samples further improve downstream performance in limited-data regimes. Beyond generation quality, the strong downstream performance of intermediate DiT representations indicates that the generative backbone learns transferable neural features, rather than merely fitting realistic signal trajectories. The advantage of BrainWorld$_{\mathrm{F+S\text{-}prior}}$ over BrainWorld$_{\mathrm{F}}$ further suggests that sMRI conditioning benefits both predictive generation and representation learning.
A key implication is that structure-function fusion can be more effective when embedded into the generative process rather than applied as post-hoc feature fusion. Unlike multimodal models such as Brain Harmony that compress sMRI and fMRI into shared representations, BrainWorld assigns sMRI a distinct role as subject-level anatomical context that modulates functional dynamics generation. Moreover, BrainWorld$_{\mathrm{F+S\text{-}prior}}$ outperforms the same-backbone BrainWorld$_{\mathrm{F+S\text{-}concat}}$ baseline, suggesting that the gain does not simply come from adding sMRI features, but from injecting anatomical information through the DiT structural-prior pathway. This design reflects a useful division of labor: the VAE preserves voxel-level spatial organization in a compact latent space, the DiT models temporal evolution, FC embeddings provide dynamic functional context during generation, and sMRI supplies stable subject-level anatomical context.

\noindent\textbf{Limitations.}
BrainWorld currently relies on next-window prediction with diffusion-based autoregressive rollout, which may accumulate errors and increase inference latency over very long horizons. 
Although the framework supports multiple conditions, our experiments evaluate condition-aware prediction rather than explicit counterfactual control; fine-grained task- and stimulus-level control also depends on the quality and temporal alignment of external embeddings. 
Moreover, generated examples are used as label-preserving augmentations rather than independent biological measurements, and the benefit of structural conditioning should be interpreted as computational evidence for useful structure-function coupling rather than biological causality. 
Future work should explore hierarchical temporal rollout, accelerated samplers or distillation, richer stimulus and task conditioning, uncertainty-aware generation, and more mechanistic validation of structure-function relationships.
\bibliographystyle{plainnat}
\bibliography{bibs/stroy,bibs/fMRI_tech,bibs/Med_tech,bibs/mode_tech,bibs/dataset}

\clearpage
\appendix
\input{supplementary}

\end{document}

%% file: supplementary.tex
\section*{Supplementary Material}
\setcounter{section}{0}
\setcounter{subsection}{0}
\setcounter{subsubsection}{0}

\renewcommand{\thesection}{\Alph{section}.}
\renewcommand{\thesubsection}{\Alph{section}.\arabic{subsection}}
\renewcommand{\thesubsubsection}{\thesubsection.\arabic{subsubsection}}

\setcounter{figure}{0}
\renewcommand{\thefigure}{S\arabic{figure}}
\renewcommand{\theHfigure}{supp.\arabic{figure}}
\setcounter{table}{0}
\renewcommand{\thetable}{S\arabic{table}}
\renewcommand{\theHtable}{supp.\arabic{table}}

\section{ROI baseline for comparison: BrainMAE}

\subsection{Architecture and Qualitative Reconstruction}
\label{supp:BrainMAE}
\begin{figure}[H]
    \centering
    \includegraphics[width=\linewidth]{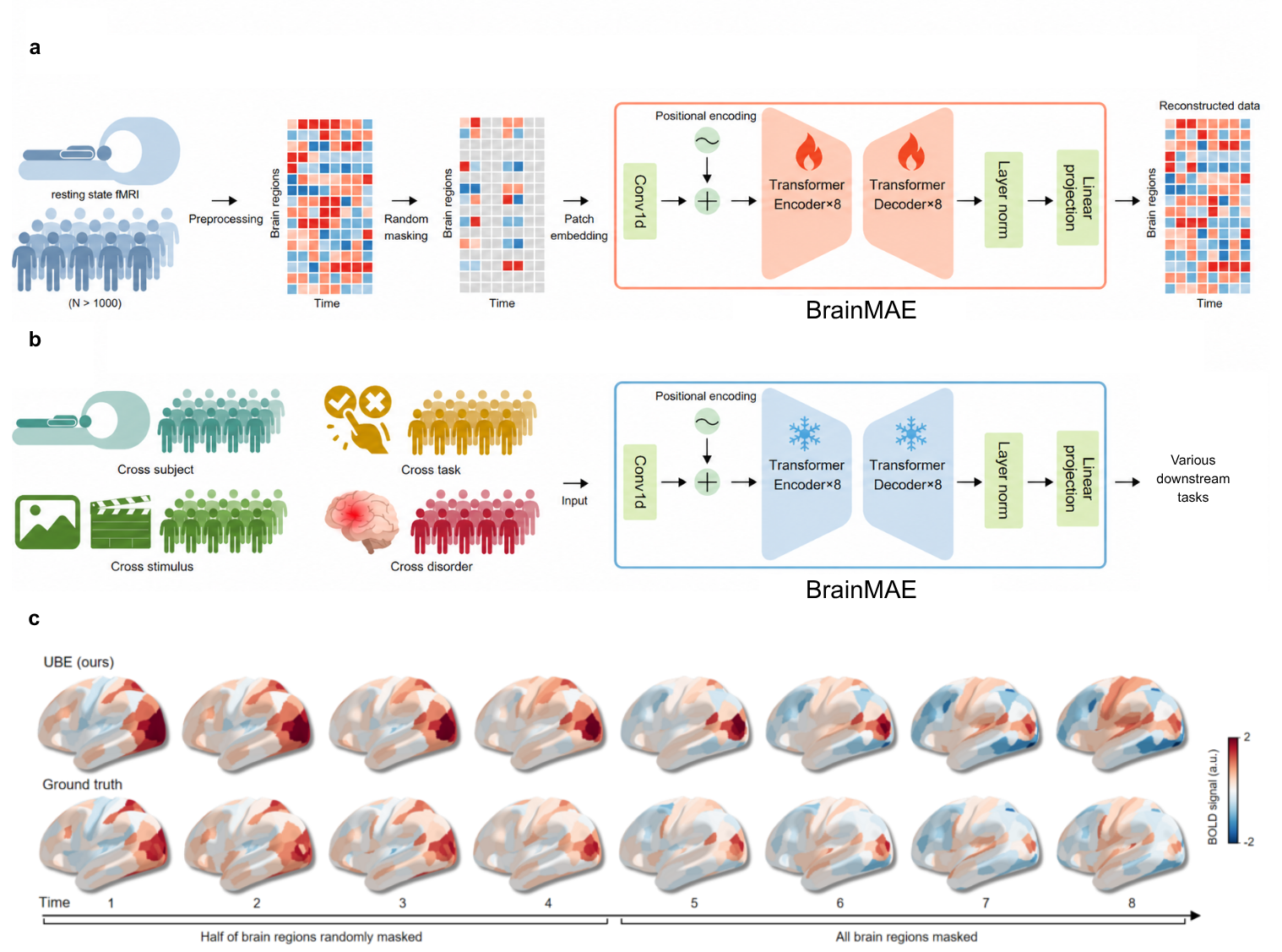}
    \caption{
    Overview of BrainMAE.
    (a) BrainMAE is pretrained on MMP-360 ROI time series using random temporal masking and full-sequence reconstruction.
    (b) After pretraining, the encoder is reused as a frozen feature extractor for probing-style transfer across subjects, tasks, naturalistic stimuli, and disease cohorts.
    (c) Qualitative reconstruction examples under half-region masking and full-region masking, showing that BrainMAE recovers temporally coherent ROI dynamics even under strong masking.
    }
    \label{fig:supp_visualization_1}
\end{figure}

To provide a controlled ROI-level comparison, we implement BrainMAE as a masked-reconstruction baseline using MMP-360 ROI time series.
It operates on MMP-360 ROI time series~\cite{MMP}, yielding inputs of shape \((B,360,T)\). 
As shown in Fig.~\ref{fig:supp_visualization_1}(a,b), a 1D convolutional patch embedding with kernel size 5 and stride 5 tokenizes the temporal axis into \(T/5\) tokens. 
BrainMAE follows a SimMIM-style objective~\cite{simmim}, where random temporal patches are masked and the full ROI--time sequence is reconstructed from visible tokens. 
The backbone is a Nystromformer encoder--decoder~\cite{nystromformer} with hidden size 1024 and 8 attention heads, trained with MSE loss over all ROI--time positions.

We train two BrainMAE variants. 
An HCP-only model is trained on HCP resting-state data to evaluate reconstruction generalization across new subjects, tasks, stimuli, and disease cohorts. 
For the main baseline comparisons, we train a larger multi-dataset model on HCP~\cite{HCP}, CHCP~\cite{CHCP}, HCP Task~\cite{HCP}, the Nakai--Nishimoto cognitive-task dataset~\cite{nakai_quantitative_2020}, HCP Movie~\cite{HCP}, StudyForrest~\cite{StudyForrest}, NSD~\cite{allen_massive_2022}, ABIDE~\cite{ABIDE}, ADHD~\cite{adhd2012adhd}, and ADNI~\cite{ADNI}. 
This model is used only as the BrainMAE baseline and is not counted in the 22-dataset BrainWorld corpus.

\subsection{Reconstruction Statistics Across Conditions and Masking Settings}

\begin{figure}[t]
    \centering
    \includegraphics[width=\linewidth]{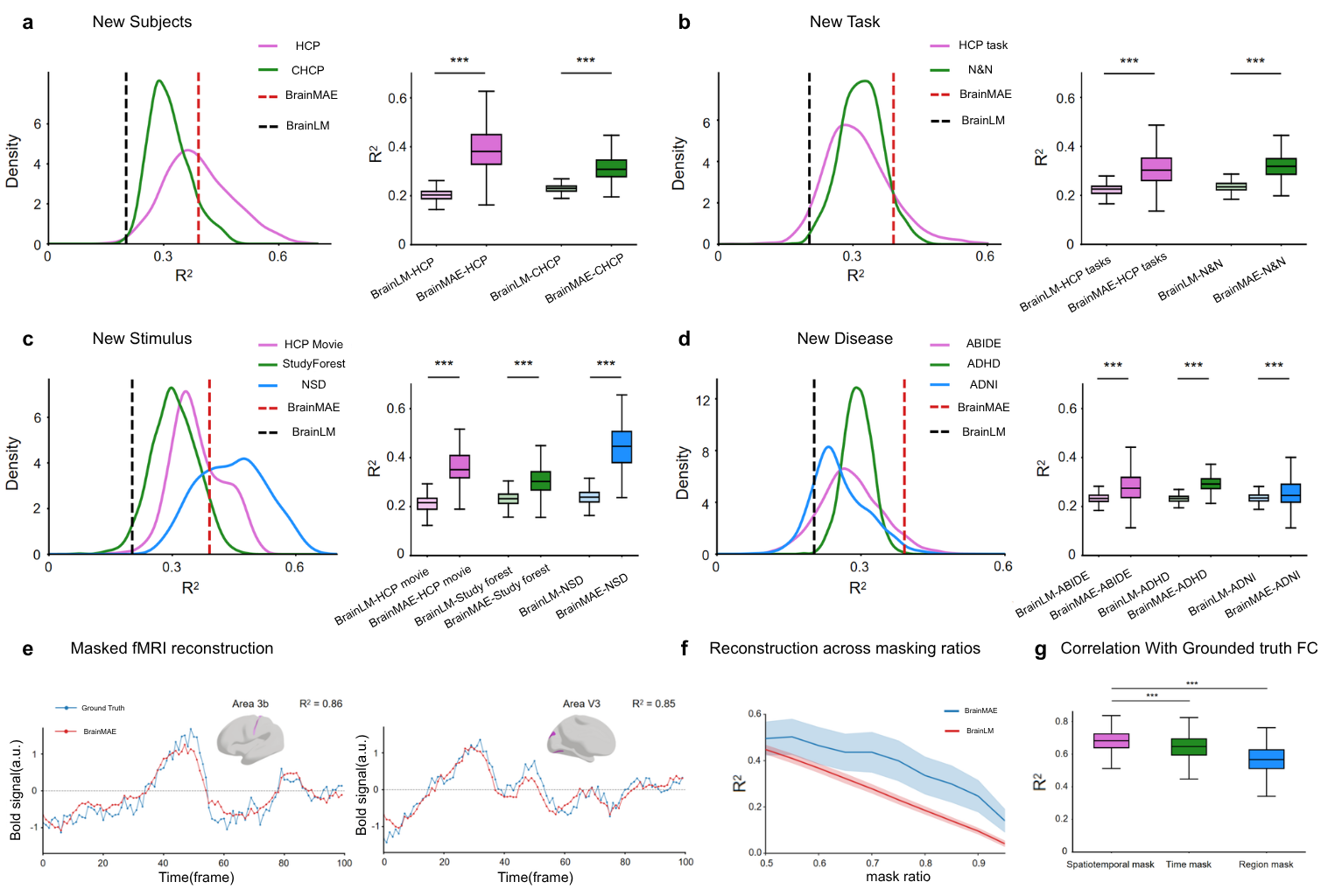}
    \caption{
        Reconstruction statistics of BrainMAE across transfer conditions and masking settings.
        (a--d) Reconstruction generalization of the HCP-only BrainMAE on new subjects, unseen tasks, naturalistic-stimulus data, and disease cohorts, compared with BrainLM.
        (e) Representative ROI-level time-course reconstruction.
        (f) Reconstruction performance across masking ratios.
        (g) Correlation between reconstructed and ground-truth functional connectivity under different masking strategies.
        }
    \label{fig:supp_visualization_2}
\end{figure}

Figure~\ref{fig:supp_visualization_2} evaluates the reconstruction generalization of the HCP-only BrainMAE across multiple transfer settings. Panels (a--d) show that, after being trained only on HCP resting-state data, BrainMAE remains competitive on new subjects, unseen tasks, naturalistic-stimulus data, and disease cohorts, consistently outperforming BrainLM under the same ROI-level reconstruction setting. Panels (e--g) further examine reconstruction behavior under different masking settings. BrainMAE recovers representative ROI time courses, remains robust across masking ratios, and preserves functional-connectivity structure. These results indicate that BrainMAE learns stable spatiotemporal dynamics, supporting its use as a strong predictive-reconstruction baseline.

\subsection{Integrated-Gradient and Connectivity Analysis}

We further analyze whether BrainMAE captures inter-regional dependencies during reconstruction using an attribution-based approach. Let \(X\in\mathbb{R}^{N\times T}\) denote an ROI--time sequence with \(N=360\) ROIs and \(T\) time points. For each target ROI \(j\), we mask its signal to obtain \(X_{\mathrm{mask}}^{(j)}\), reconstruct it with the pretrained BrainMAE \(f_\theta\), and compute integrated gradients from a constant baseline \(X_0\) to the masked input~\cite{intengrated}. The interpolation path is defined as
\[
X_{\alpha}^{(j)}
=
X_0+\alpha\left(X_{\mathrm{mask}}^{(j)}-X_0\right),
\quad \alpha\in[0,1].
\]
The target reconstruction loss is
\[
\mathcal{L}_{j}(\alpha)
=
\left\|
f_\theta\!\left(X_{\alpha}^{(j)}\right)_{j,:}
-
X_{j,:}
\right\|_2^2 .
\]
For source ROI \(i\), we define its attribution to target ROI \(j\) as
\[
A_{i\rightarrow j}
=
\frac{1}{T}
\sum_{t=1}^{T}
\left|
X_{\mathrm{mask},i,t}^{(j)}-X_{0,i,t}
\right|
\cdot
\frac{1}{K}
\sum_{k=1}^{K}
\left|
\frac{
\partial \mathcal{L}_{j}(\alpha_k)
}{
\partial X_{\alpha_k,i,t}^{(j)}
}
\right|,
\quad
\alpha_k=\frac{k-1}{K-1},
\]
where \(K=50\) interpolation steps are used. The resulting attributions form a directed gradient-connectivity matrix \(GC\in\mathbb{R}^{N\times N}\), with \(GC_{i,j}=A_{i\rightarrow j}\), and diagonal entries set to zero. Larger \(GC_{i,j}\) values indicate that source ROI \(i\) has a stronger influence on reconstructing the masked target ROI \(j\).

\begin{figure}[t]
    \centering
    \includegraphics[width=\linewidth]{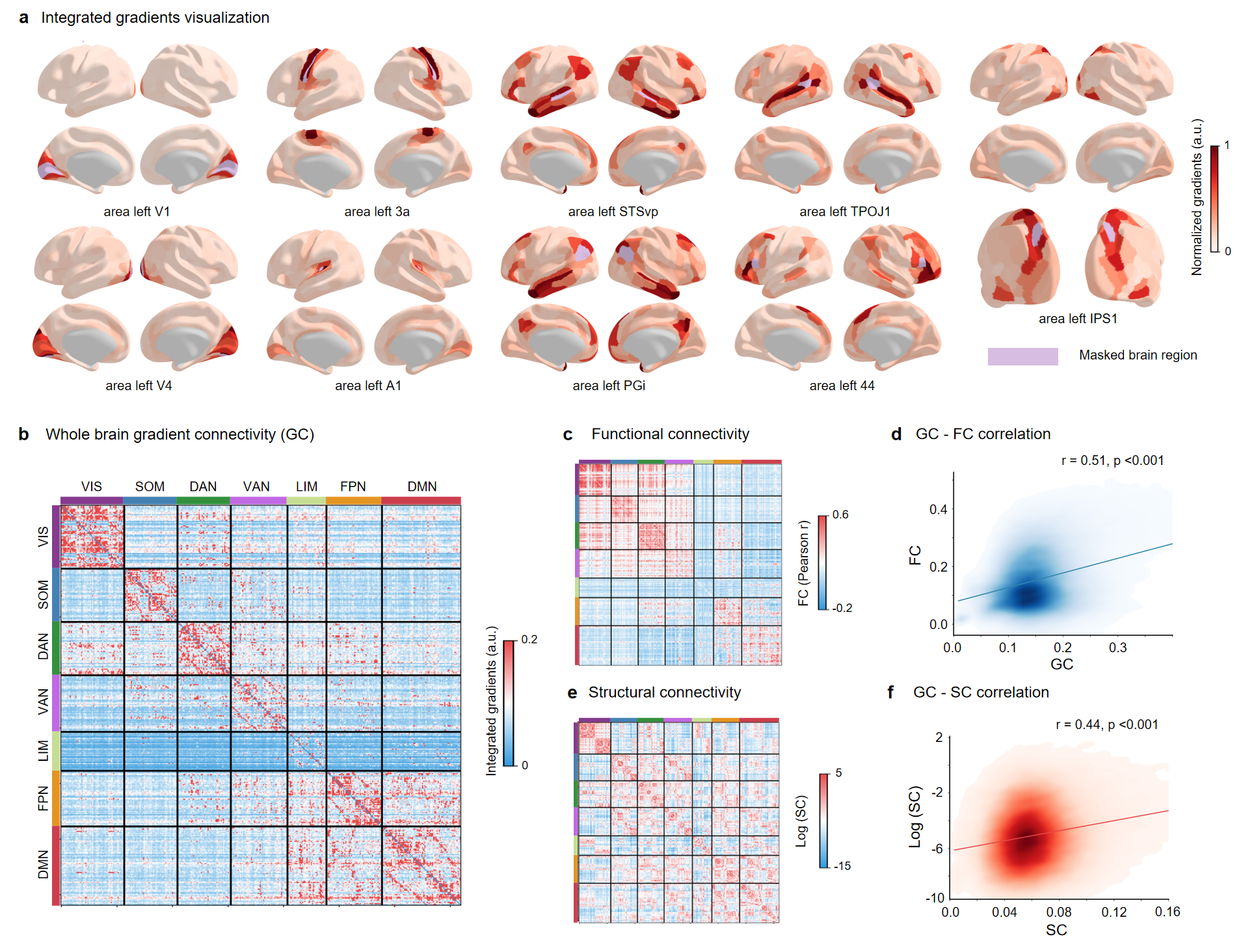}
    \caption{
    Integrated-gradient and connectivity analysis of BrainMAE.
    (a) Integrated-gradient attribution maps for representative masked target ROIs.
    (b) Whole-brain gradient-connectivity (GC) matrix aggregated across all 360 ROIs.
    (c) Functional connectivity (FC) matrix.
    (d) Correlation between GC and FC.
    (e) Structural connectivity (SC) matrix.
    (f) Correlation between GC and SC.
    Together, these results suggest that BrainMAE learns distributed inter-regional dependencies consistent with functional and structural brain organization during reconstruction.
    }
    \label{fig:supp_visualization_3}
\end{figure}

Figure~\ref{fig:supp_visualization_3} visualizes the resulting attribution and connectivity patterns. Panel (a) shows representative integrated-gradient maps, indicating that reconstruction depends on distributed and anatomically structured patterns rather than purely local information. Panel (b) aggregates these attributions across all 360 ROIs to form a whole-brain gradient-connectivity matrix. Panels (c--f) compare GC against functional connectivity (FC) and structural connectivity (SC). GC is positively correlated with both FC (\(r=0.51, p<0.001\)) and SC (\(r=0.44, p<0.001\)), suggesting that BrainMAE learns inter-regional dependencies consistent with functional and structural brain organization during reconstruction.

\subsection{Downstream Transfer Evaluation}

\begin{figure}[t]
    \centering
    \includegraphics[width=\linewidth]{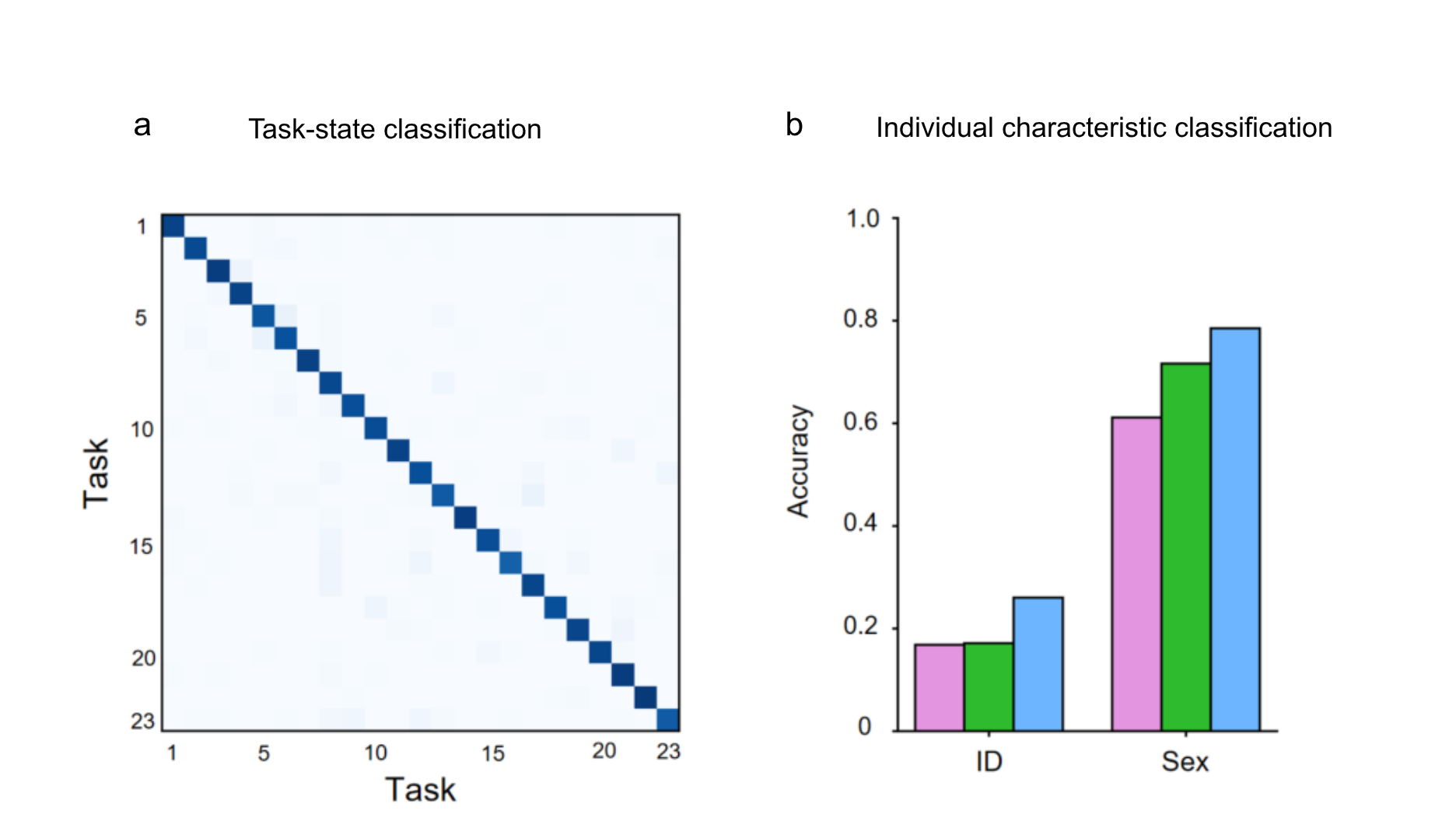}
    \caption{
    Downstream transfer evaluation of BrainMAE.
    (a) HCP task-state classification across 23 task categories, showing strong separation of task-specific representations.
    (b) Individual-level characteristic prediction, including subject identification and sex classification, compared with simpler representation baselines.
    }
    \label{fig:supp_visualization_4}
\end{figure}

We further evaluate whether BrainMAE learns representations that transfer beyond masked reconstruction. The pretrained BrainMAE encoder is frozen and used as a feature extractor, followed by lightweight downstream classifiers. Figure~\ref{fig:supp_visualization_4}(a) shows HCP task-state classification across 23 task categories. The near-diagonal confusion matrix indicates that BrainMAE representations preserve strong task-discriminative information. Figure~\ref{fig:supp_visualization_4}(b) further evaluates individual-level characteristics, including subject identification and sex classification, where BrainMAE achieves higher accuracy than simpler representation baselines. These results show that BrainMAE is not only a reconstruction baseline, but also a competitive transfer baseline with meaningful task- and subject-level representations.

\section{Supplementary Results}

\subsection{Additional 4D fMRI Generation Visualizations}
\label{supp:Generation_Visualizations}

\begin{figure}[H]
    \centering
    \includegraphics[width=\linewidth]{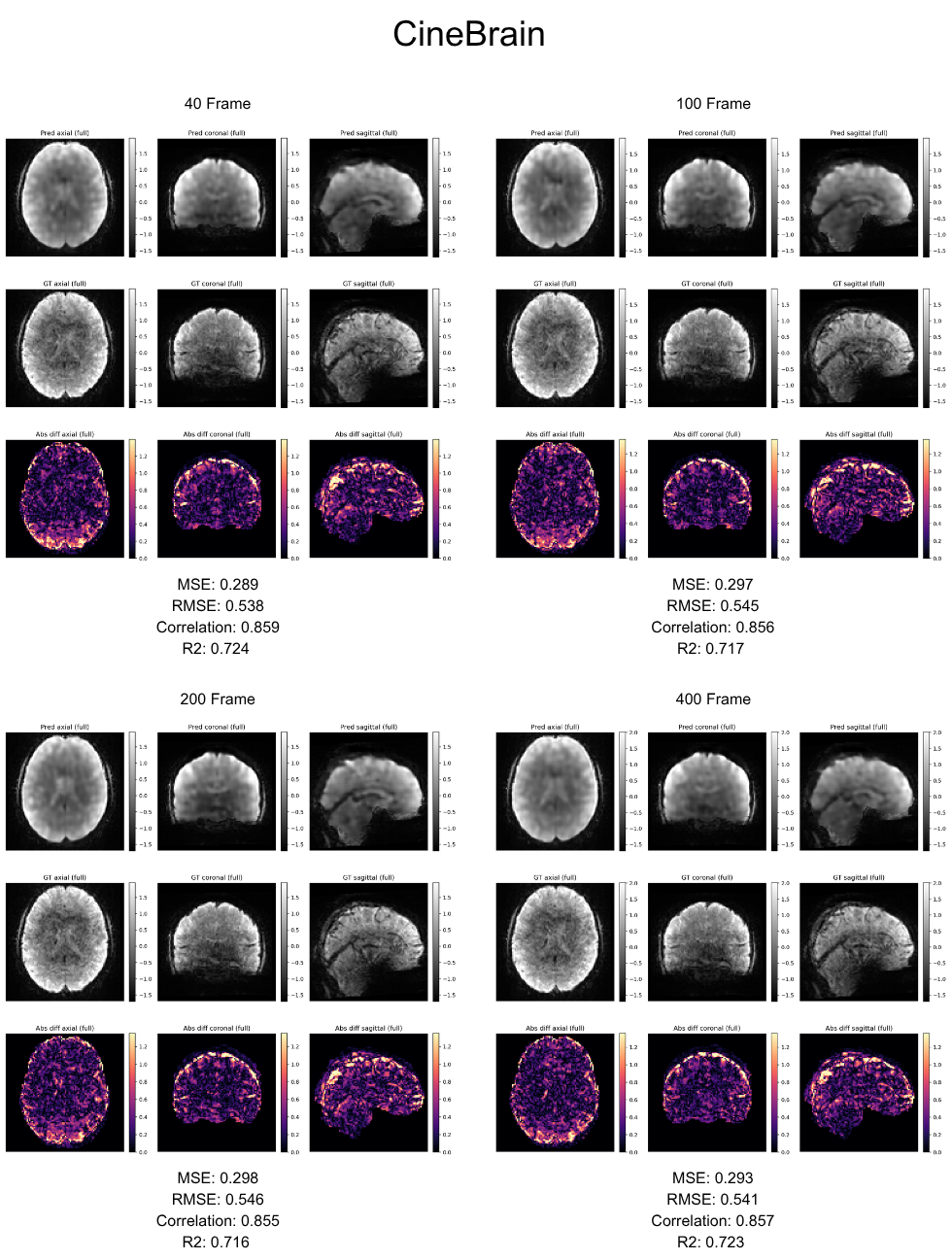}
    \caption{
    Qualitative visualization of BrainWorld generation on CineBrain across prediction horizons.
    For each horizon, axial, coronal, and sagittal slices are shown for generated fMRI, ground truth, and absolute error maps.
    The displayed subject/run is selected as the median-performing case according to MSE among all evaluated CineBrain samples, rather than as the best case.
    Metrics below each panel report MSE, RMSE, Pearson correlation, and \(R^2\).
    }
    \label{fig:supp_CineBrain_visual}
\end{figure}

\begin{figure}[H]
    \centering
    \includegraphics[width=\linewidth]{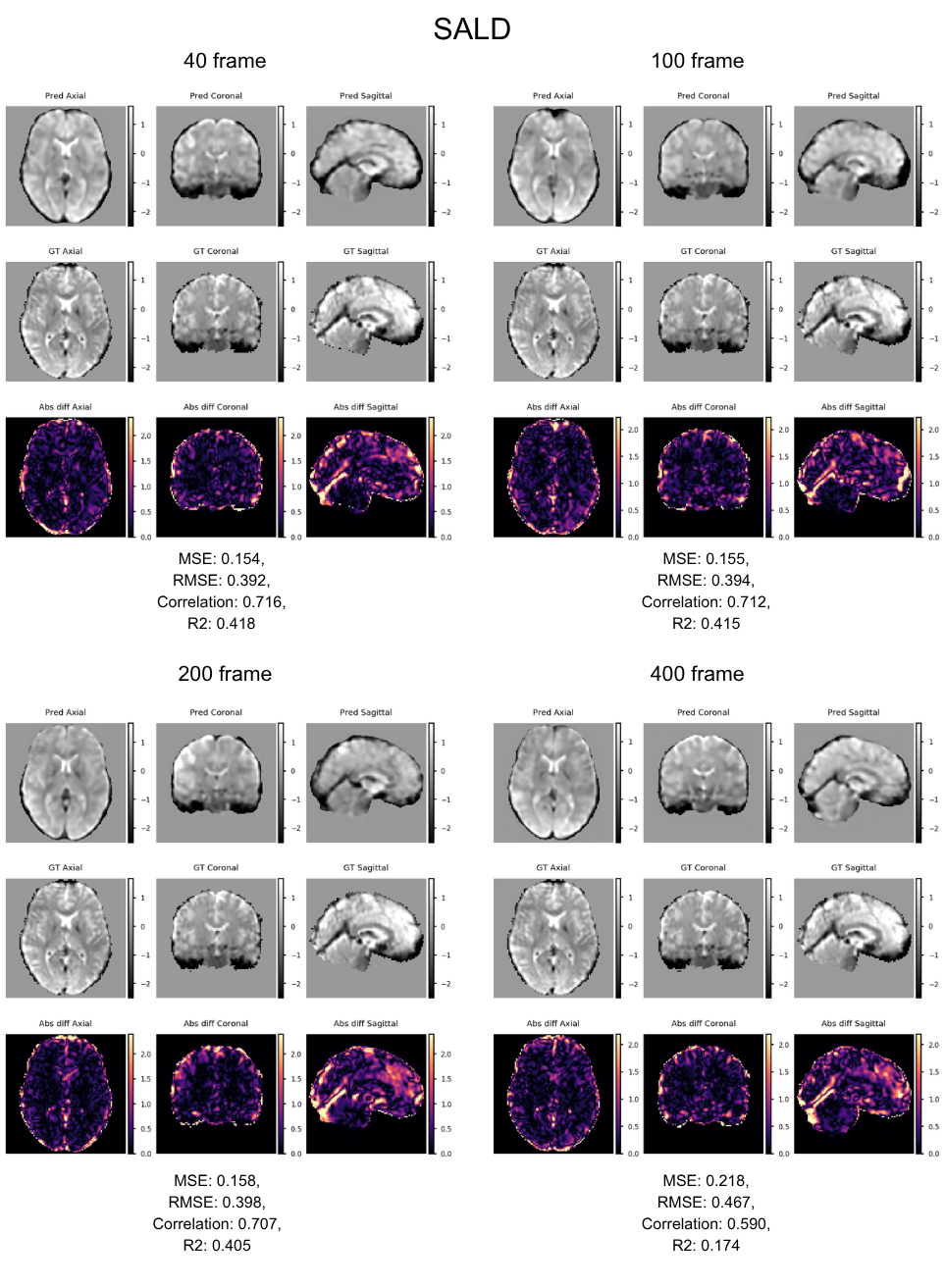}
    \caption{
    Qualitative visualization of BrainWorld generation on SALD across prediction horizons.
    For each horizon, axial, coronal, and sagittal slices are shown for generated fMRI, ground truth, and absolute error maps.
    The displayed subject is selected as the median-performing case according to MSE among all evaluated SALD samples, rather than as the best case.
    Metrics below each panel report MSE, RMSE, Pearson correlation, and \(R^2\).
    }
    \label{fig:supp_Sald_visual}
\end{figure}

\begin{figure}[H]
    \centering
    \includegraphics[width=\linewidth]{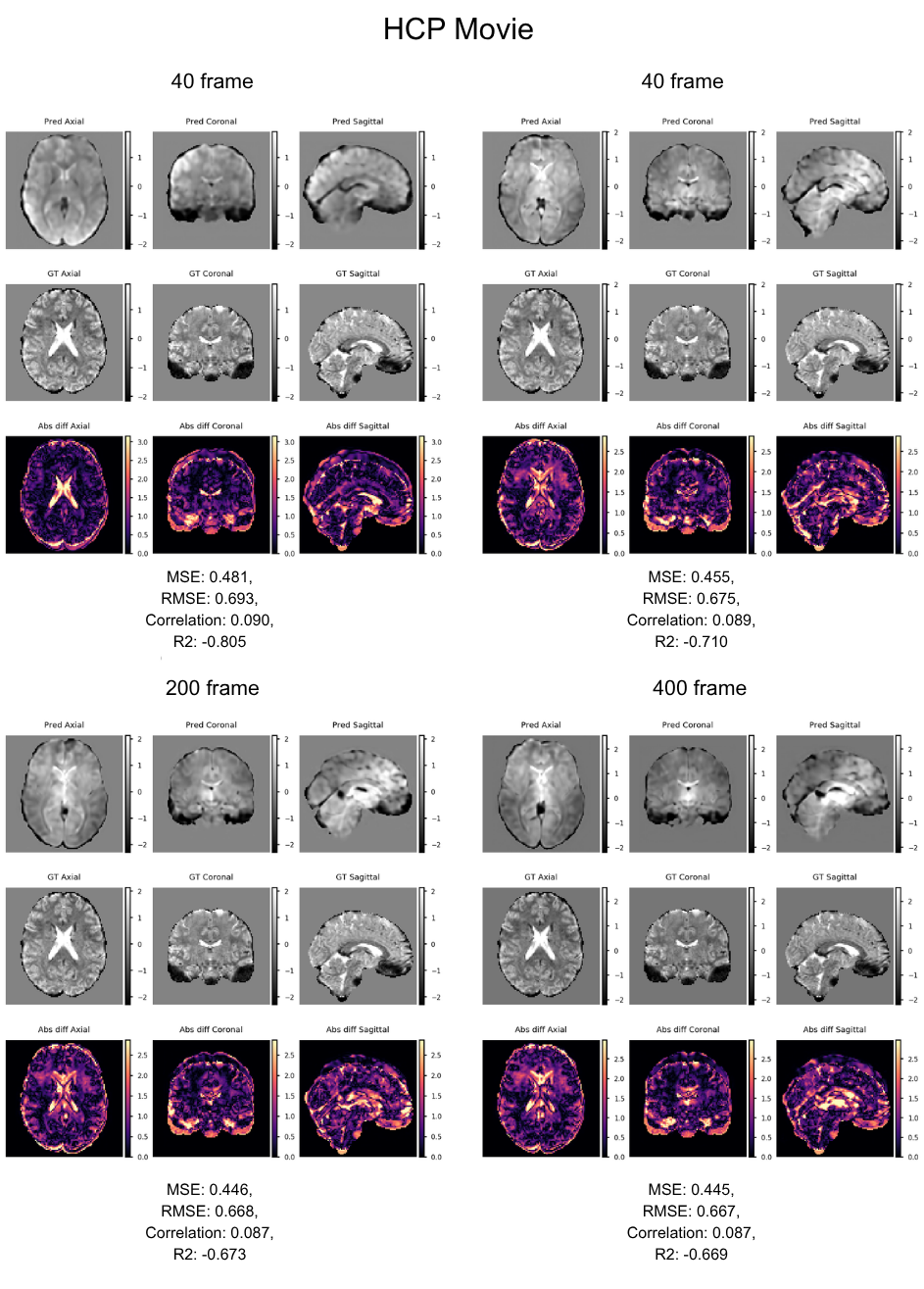}
    \caption{
    Qualitative visualization of BrainWorld generation on HCP Movie across prediction horizons.
    For each horizon, axial, coronal, and sagittal slices are shown for generated fMRI, ground truth, and absolute error maps.
    The displayed run is selected as the median-performing case according to MSE among all evaluated HCP Movie runs, rather than as the best case.
    Metrics below each panel report MSE, RMSE, Pearson correlation, and \(R^2\).
    }
    \label{fig:supp_HCP_Movie_visual}
\end{figure}

To complement the quantitative generation results, we provide qualitative visualizations of BrainWorld predictions on three representative datasets: CineBrain (Fig.~\ref{fig:supp_CineBrain_visual}), SALD (Fig.~\ref{fig:supp_Sald_visual}), and HCP Movie (Fig.~\ref{fig:supp_HCP_Movie_visual}).
For each dataset, we visualize predictions at multiple rollout horizons using axial, coronal, and sagittal slices, together with the corresponding ground truth and absolute error maps. 
To avoid cherry-picking visually favorable examples, the displayed subject or run is selected as the median-performing case according to MSE among all evaluated samples of the corresponding dataset. 
Across CineBrain and SALD, BrainWorld preserves coherent whole-brain spatial structure over increasing horizons, while the HCP Movie results illustrate the greater difficulty of external naturalistic-stimulus prediction.

\subsection{Generation Statistics at Additional Rollout Lengths}
\label{supp:Generation_Statistics}
We report quantitative generation performance at four rollout lengths: 40, 100, 200, and 400 frames (Tables~\ref{tab:supp_generation_40}--\ref{tab:supp_generation_400}). 
For each horizon, we evaluate generation quality on HCP LR1, SALD, CineBrain, and HCP Movie using MSE, RMSE, and Pearson correlation. 
Lower MSE/RMSE and higher correlation indicate better reconstruction of future fMRI dynamics. 
Across rollout lengths, BrainWorld$_{\mathrm{F+S\text{-}prior}}$ maintains stable performance and achieves the best or near-best results on most external evaluation settings, especially on SALD and HCP Movie.
The results also show that performance generally degrades as the prediction horizon increases, reflecting the increasing difficulty of long-horizon autoregressive generation.

\begin{table*}[t]
\centering 
\small
\setlength{\tabcolsep}{2pt} 
\renewcommand{\arraystretch}{1.15}
\caption{
Generation performance for 40-frame prediction across HCP, SALD, CineBrain, and HCP Movie.
MSE, RMSE, and Pearson correlation are reported. Lower MSE/RMSE and higher correlation indicate better performance. \best{Red} indicates the best performance and \underline{Underline} indicates the second-best performance.
\(^{*}\) denotes a large favorable effect size with Cohen's \(d \ge 0.8\) compared with the best-performing non-BrainWorld baseline for the same dataset, horizon, and metric.
}
\label{tab:supp_generation_40}

\resizebox{\textwidth}{!}{
\begin{tabular}{l ccc ccc}
\toprule 
\multirow{3}{*}{\diagbox[width=7em]{\textbf{Model}}{\textbf{Dataset}}} &
\multicolumn{3}{c}{\textbf{HCP LR1}} & \multicolumn{3}{c}{\textbf{SALD}} \\
& \multicolumn{3}{c}{\textit{Internal}} & \multicolumn{3}{c}{\textit{External}} \\
\cmidrule(lr){2-4} \cmidrule(lr){5-7}
& \textbf{MSE $\downarrow$} & \textbf{R $\uparrow$} & \textbf{RMSE $\downarrow$} & \textbf{MSE $\downarrow$} & \textbf{R $\uparrow$} & \textbf{RMSE $\downarrow$} \\
\midrule
BrainMAE & 0.986$\pm$0.126 & 0.147$\pm$0.140 & 0.991$\pm$0.062 & 1.038$\pm$0.097 & 0.055$\pm$0.181 & 1.018$\pm$0.047 \\
BrainLM & 0.925$\pm$0.084 & 0.265$\pm$0.087 & 0.961$\pm$0.041 & 0.573$\pm$0.292 & 0.429$\pm$0.156 & 0.738$\pm$0.171 \\
NeuroSTORM & 0.514$\pm$0.020 & 0.182$\pm$0.033 & 0.716$\pm$0.014 & 0.590$\pm$0.023 & 0.170$\pm$0.019 & 0.768$\pm$0.015 \\
MONAI & \best{\textbf{0.157$\pm$0.014}} & \best{\textbf{0.666$\pm$0.032}} & \best{\textbf{0.396$\pm$0.017}} & \underline{0.269$\pm$0.030} & \underline{0.496$\pm$0.069} & \underline{0.516$\pm$0.029} \\
\midrule
BrainWorld & \underline{0.175$\pm$0.027} & \underline{0.659$\pm$0.052} & \underline{0.417$\pm$0.029} & \best{\textbf{0.186$\pm$0.074*}} & \best{\textbf{0.657$\pm$0.140*}} & \best{\textbf{0.425$\pm$0.074*}} \\
\bottomrule
\end{tabular}
}

\vspace{0.8em}

\resizebox{\textwidth}{!}{
\begin{tabular}{l ccc ccc}
\toprule 
\multirow{3}{*}{\diagbox[width=7em]{\textbf{Model}}{\textbf{Dataset}}} &
\multicolumn{3}{c}{\textbf{CineBrain}} & \multicolumn{3}{c}{\textbf{HCP Movie}} \\
& \multicolumn{3}{c}{\textit{Internal}} & \multicolumn{3}{c}{\textit{External}} \\
\cmidrule(lr){2-4} \cmidrule(lr){5-7}
& \textbf{MSE $\downarrow$} & \textbf{R $\uparrow$} & \textbf{RMSE $\downarrow$} & \textbf{MSE $\downarrow$} & \textbf{R $\uparrow$} & \textbf{RMSE $\downarrow$} \\
\midrule
BrainMAE & 0.781$\pm$0.083 & 0.456$\pm$0.107 & 0.882$\pm$0.048 & 1.001$\pm$0.031 & 0.075$\pm$0.083 & 1.000$\pm$0.016 \\
BrainLM & 0.766$\pm$0.407 & 0.522$\pm$0.128 & 0.850$\pm$0.214 & 1.059$\pm$0.162 & \underline{0.133$\pm$0.069} & 1.027$\pm$0.071 \\
NeuroSTORM & 0.795$\pm$0.034 & 0.628$\pm$0.004 & 0.891$\pm$0.019 & \underline{0.517$\pm$0.008} & 0.082$\pm$0.009 & \underline{0.719$\pm$0.006} \\
MONAI & \underline{0.422$\pm$0.022} & \underline{0.752$\pm$0.016} & \underline{0.649$\pm$0.017} & 0.871$\pm$0.756 & 0.056$\pm$0.041 & 0.857$\pm$0.756 \\
\midrule
BrainWorld & \best{\textbf{0.292$\pm$0.021}$^{*}$} & \best{\textbf{0.859$\pm$0.009}$^{*}$} & \best{\textbf{0.540$\pm$0.019}$^{*}$} & \best{\textbf{0.405$\pm$0.095}$^{*}$} & \best{\textbf{0.180$\pm$0.152}} & \best{\textbf{0.631$\pm$0.077}$^{*}$} \\
\bottomrule
\end{tabular}
}

\end{table*}

\begin{table*}[t]
\centering 
\small
\setlength{\tabcolsep}{2pt} 
\renewcommand{\arraystretch}{1.15}
\caption{
Generation performance for 100-frame prediction across HCP, SALD, CineBrain, and HCP Movie.
MSE, RMSE, and Pearson correlation are reported. Lower MSE/RMSE and higher correlation indicate better performance. \best{Red} indicates the best performance and \underline{Underline} indicates the second-best performance.
* denotes a large favorable effect size with Cohen's \(d \ge 0.8\) compared with the best-performing non-BrainWorld baseline for the same dataset, horizon, and metric.
}
\label{tab:supp_generation_100}

\resizebox{\textwidth}{!}{
\begin{tabular}{l ccc ccc}
\toprule 
\multirow{3}{*}{\diagbox[width=7em]{\textbf{Model}}{\textbf{Dataset}}} &
\multicolumn{3}{c}{\textbf{HCP LR1}} & \multicolumn{3}{c}{\textbf{SALD}} \\
& \multicolumn{3}{c}{\textit{Internal}} & \multicolumn{3}{c}{\textit{External}} \\
\cmidrule(lr){2-4} \cmidrule(lr){5-7}
& \textbf{MSE $\downarrow$} & \textbf{R $\uparrow$} & \textbf{RMSE $\downarrow$} & \textbf{MSE $\downarrow$} & \textbf{R $\uparrow$} & \textbf{RMSE $\downarrow$} \\
\midrule
BrainMAE & 1.001$\pm$0.071 & 0.096$\pm$0.102 & 1.000$\pm$0.035 & 1.018$\pm$0.055 & 0.033$\pm$0.122 & 1.009$\pm$0.027 \\
BrainLM & 0.971$\pm$0.032 & 0.171$\pm$0.059 & 0.985$\pm$0.016 & 0.724$\pm$0.284 & 0.331$\pm$0.148 & 0.836$\pm$0.161 \\
NeuroSTORM & 0.944$\pm$0.038 & 0.079$\pm$0.022 & 0.971$\pm$0.019 & 0.991$\pm$0.032 & 0.072$\pm$0.017 & 0.996$\pm$0.016 \\
MONAI & \underline{0.318$\pm$0.017} & \underline{0.394$\pm$0.030} & \underline{0.564$\pm$0.015} & \underline{0.283$\pm$0.017} & \underline{0.510$\pm$0.036} & \underline{0.532$\pm$0.016} \\
\midrule
BrainWorld & \best{\textbf{0.181$\pm$0.026*}} & \best{\textbf{0.645$\pm$0.050*}} & \best{\textbf{0.425$\pm$0.028*}} & \best{\textbf{0.204$\pm$0.078*}} & \best{\textbf{0.623$\pm$0.146*}} & \best{\textbf{0.445$\pm$0.077*}} \\
\bottomrule
\end{tabular}
}

\vspace{0.8em}

\resizebox{\textwidth}{!}{
\begin{tabular}{l ccc ccc}
\toprule 
\multirow{3}{*}{\diagbox[width=7em]{\textbf{Model}}{\textbf{Dataset}}} &
\multicolumn{3}{c}{\textbf{CineBrain}} & \multicolumn{3}{c}{\textbf{HCP Movie}} \\
& \multicolumn{3}{c}{\textit{Internal}} & \multicolumn{3}{c}{\textit{External}} \\
\cmidrule(lr){2-4} \cmidrule(lr){5-7}
& \textbf{MSE $\downarrow$} & \textbf{R $\uparrow$} & \textbf{RMSE $\downarrow$} & \textbf{MSE $\downarrow$} & \textbf{R $\uparrow$} & \textbf{RMSE $\downarrow$} \\
\midrule
BrainMAE & 0.908$\pm$0.068 & 0.299$\pm$0.077 & 0.952$\pm$0.037 & 1.003$\pm$0.018 & 0.042$\pm$0.055 & 1.001$\pm$0.009 \\
BrainLM & 0.840$\pm$0.574 & 0.473$\pm$0.123 & 0.878$\pm$0.271 & 1.060$\pm$0.138 & \underline{0.077$\pm$0.042} & 1.028$\pm$0.062 \\
NeuroSTORM & 1.182$\pm$0.040 & 0.368$\pm$0.006 & 1.087$\pm$0.019 & 0.932$\pm$0.017 & 0.023$\pm$0.005 & 0.965$\pm$0.009 \\
MONAI & \underline{0.684$\pm$0.027} & \underline{0.549$\pm$0.020} & \underline{0.827$\pm$0.017} & \underline{0.595$\pm$0.322} & 0.046$\pm$0.034 & \underline{0.747$\pm$0.191} \\
\midrule
BrainWorld & \best{\textbf{0.291$\pm$0.021*}} & \best{\textbf{0.860$\pm$0.010*}} & \best{\textbf{0.539$\pm$0.019*}} & \best{\textbf{0.437$\pm$0.045}} & \best{\textbf{0.114$\pm$0.072}} & \best{\textbf{0.660$\pm$0.035}} \\
\bottomrule
\end{tabular}
}

\end{table*}

\begin{table*}[t]
\centering 
\small
\setlength{\tabcolsep}{2pt} 
\renewcommand{\arraystretch}{1.15}
\caption{
Generation performance for 200-frame prediction across HCP, SALD, CineBrain, and HCP Movie.
MSE, RMSE, and Pearson correlation are reported. Lower MSE/RMSE and higher correlation indicate better performance. \best{Red} indicates the best performance and \underline{Underline} indicates the second-best performance.
* denotes a large favorable effect size with Cohen's \(d \ge 0.8\) compared with the best-performing non-BrainWorld baseline for the same dataset, horizon, and metric.
}
\label{tab:supp_generation_200}

\resizebox{\textwidth}{!}{
\begin{tabular}{l ccc ccc}
\toprule 
\multirow{3}{*}{\diagbox[width=7em]{\textbf{Model}}{\textbf{Dataset}}} &
\multicolumn{3}{c}{\textbf{HCP LR1}} & \multicolumn{3}{c}{\textbf{SALD}} \\
& \multicolumn{3}{c}{\textit{Internal}} & \multicolumn{3}{c}{\textit{External}} \\
\cmidrule(lr){2-4} \cmidrule(lr){5-7}
& \textbf{MSE $\downarrow$} & \textbf{R $\uparrow$} & \textbf{RMSE $\downarrow$} & \textbf{MSE $\downarrow$} & \textbf{R $\uparrow$} & \textbf{RMSE $\downarrow$} \\
\midrule
BrainMAE & 0.999$\pm$0.048 & 0.066$\pm$0.073 & 0.999$\pm$0.024 & 1.010$\pm$0.023 & 0.021$\pm$0.087 & 1.005$\pm$0.012 \\
BrainLM & 0.988$\pm$0.019 & 0.119$\pm$0.043 & 0.994$\pm$0.010 & 0.889$\pm$0.303 & 0.229$\pm$0.148 & 0.929$\pm$0.158 \\
NeuroSTORM & 1.168$\pm$0.041 & 0.038$\pm$0.019 & 1.080$\pm$0.019 & 1.164$\pm$0.041 & 0.030$\pm$0.013 & 1.079$\pm$0.019 \\
MONAI & \underline{0.375$\pm$0.021} & \underline{0.304$\pm$0.036} & \underline{0.613$\pm$0.017} & \underline{0.287$\pm$0.015} & \underline{0.513$\pm$0.030} & \underline{0.536$\pm$0.014} \\
\midrule
BrainWorld & \best{\textbf{0.183$\pm$0.027*}} & \best{\textbf{0.640$\pm$0.051*}} & \best{\textbf{0.427$\pm$0.028*}} & \best{\textbf{0.214$\pm$0.081*}} & \best{\textbf{0.604$\pm$0.153*}} & \best{\textbf{0.455$\pm$0.080*}} \\
\bottomrule
\end{tabular}
}

\vspace{0.8em}

\resizebox{\textwidth}{!}{
\begin{tabular}{l ccc ccc}
\toprule 
\multirow{3}{*}{\diagbox[width=7em]{\textbf{Model}}{\textbf{Dataset}}} &
\multicolumn{3}{c}{\textbf{CineBrain}} & \multicolumn{3}{c}{\textbf{HCP Movie}} \\
& \multicolumn{3}{c}{\textit{Internal}} & \multicolumn{3}{c}{\textit{External}} \\
\cmidrule(lr){2-4} \cmidrule(lr){5-7}
& \textbf{MSE $\downarrow$} & \textbf{R $\uparrow$} & \textbf{RMSE $\downarrow$} & \textbf{MSE $\downarrow$} & \textbf{R $\uparrow$} & \textbf{RMSE $\downarrow$} \\
\midrule
BrainMAE & 0.983$\pm$0.027 & 0.179$\pm$0.050 & 0.992$\pm$0.014 & 0.997$\pm$0.007 & 0.029$\pm$0.040 & 0.998$\pm$0.004 \\
BrainLM & \underline{0.793$\pm$0.390} & \underline{0.395$\pm$0.122} & \underline{0.867$\pm$0.209} & 1.061$\pm$0.162 & \underline{0.053$\pm$0.029} & 1.028$\pm$0.071 \\
NeuroSTORM & 1.482$\pm$0.043 & 0.176$\pm$0.005 & 1.217$\pm$0.018 & 1.157$\pm$0.025 & 0.002$\pm$0.005 & 1.076$\pm$0.011 \\
MONAI & 0.876$\pm$0.023 & 0.383$\pm$0.018 & 0.936$\pm$0.013 & \underline{0.578$\pm$0.327} & 0.051$\pm$0.033 & \underline{0.738$\pm$0.184} \\
\midrule
BrainWorld & \best{\textbf{0.298$\pm$0.029*}} & \best{\textbf{0.856$\pm$0.014*}} & \best{\textbf{0.545$\pm$0.026*}} & \best{\textbf{0.448$\pm$0.027}} & \best{\textbf{0.092$\pm$0.046*}} & \best{\textbf{0.669$\pm$0.021}} \\
\bottomrule
\end{tabular}
}

\end{table*}

\begin{table*}[t]
\centering 
\small
\setlength{\tabcolsep}{2pt} 
\renewcommand{\arraystretch}{1.15}
\caption{
Generation performance for 400-frame prediction across HCP, SALD, CineBrain, and HCP Movie.
MSE, RMSE, and Pearson correlation are reported. Lower MSE/RMSE and higher correlation indicate better performance. \best{Red} indicates the best performance and \underline{Underline} indicates the second-best performance.
* denotes a large favorable effect size with Cohen's \(d \ge 0.8\) compared with the best-performing non-BrainWorld baseline for the same dataset, horizon, and metric.
}
\label{tab:supp_generation_400}

\resizebox{\textwidth}{!}{
\begin{tabular}{l ccc ccc}
\toprule 
\multirow{3}{*}{\diagbox[width=7em]{\textbf{Model}}{\textbf{Dataset}}} &
\multicolumn{3}{c}{\textbf{HCP LR1}} & \multicolumn{3}{c}{\textbf{SALD}} \\
& \multicolumn{3}{c}{\textit{Internal}} & \multicolumn{3}{c}{\textit{External}} \\
\cmidrule(lr){2-4} \cmidrule(lr){5-7}
& \textbf{MSE $\downarrow$} & \textbf{R $\uparrow$} & \textbf{RMSE $\downarrow$} & \textbf{MSE $\downarrow$} & \textbf{R $\uparrow$} & \textbf{RMSE $\downarrow$} \\
\midrule
BrainMAE & 0.998$\pm$0.023 & 0.043$\pm$0.055 & 0.999$\pm$0.012 & 1.006$\pm$0.012 & 0.013$\pm$0.060 & 1.003$\pm$0.006 \\
BrainLM & 0.995$\pm$0.010 & 0.081$\pm$0.032 & 0.997$\pm$0.005 & 0.951$\pm$0.180 & 0.142$\pm$0.175 & 0.971$\pm$0.093 \\
NeuroSTORM & 1.208$\pm$0.049 & 0.013$\pm$0.015 & 1.099$\pm$0.022 & 1.270$\pm$0.041 & 0.005$\pm$0.011 & 1.127$\pm$0.018 \\
MONAI & \underline{0.405$\pm$0.024} & \underline{0.259$\pm$0.040} & \underline{0.636$\pm$0.019} & \underline{0.295$\pm$0.015} & \underline{0.508$\pm$0.032} & \underline{0.537$\pm$0.014} \\
\midrule
BrainWorld & \best{\textbf{0.185$\pm$0.027*}} & \best{\textbf{0.637$\pm$0.052*}} & \best{\textbf{0.429$\pm$0.028*}} & \best{\textbf{0.218$\pm$0.082*}} & \best{\textbf{0.596$\pm$0.155}} & \best{\textbf{0.460$\pm$0.081*}} \\
\bottomrule
\end{tabular}
}

\vspace{0.8em}

\resizebox{\textwidth}{!}{
\begin{tabular}{l ccc ccc}
\toprule 
\multirow{3}{*}{\diagbox[width=7em]{\textbf{Model}}{\textbf{Dataset}}} &
\multicolumn{3}{c}{\textbf{CineBrain}} & \multicolumn{3}{c}{\textbf{HCP Movie}} \\
& \multicolumn{3}{c}{\textit{Internal}} & \multicolumn{3}{c}{\textit{External}} \\
\cmidrule(lr){2-4} \cmidrule(lr){5-7}
& \textbf{MSE $\downarrow$} & \textbf{R $\uparrow$} & \textbf{RMSE $\downarrow$} & \textbf{MSE $\downarrow$} & \textbf{R $\uparrow$} & \textbf{RMSE $\downarrow$} \\
\midrule
BrainMAE & 1.006$\pm$0.014 & 0.113$\pm$0.031 & 1.003$\pm$0.007 & 0.996$\pm$0.004 & 0.020$\pm$0.028 & 0.998$\pm$0.002 \\
BrainLM & \underline{0.790$\pm$0.256} & \underline{0.297$\pm$0.106} & \underline{0.877$\pm$0.148} & 1.059$\pm$0.164 & 0.037$\pm$0.021 & 1.027$\pm$0.072 \\
NeuroSTORM & 1.653$\pm$0.048 & 0.078$\pm$0.004 & 1.286$\pm$0.019 & 1.230$\pm$0.038 & -0.008$\pm$0.006 & 1.109$\pm$0.017 \\
MONAI & 0.982$\pm$0.013 & 0.283$\pm$0.010 & 0.991$\pm$0.006 & \underline{0.528$\pm$0.230} & \underline{0.056$\pm$0.033} & \underline{0.713$\pm$0.139} \\
\midrule
BrainWorld & \best{\textbf{0.295$\pm$0.023*}} & \best{\textbf{0.857$\pm$0.011*}} & \best{\textbf{0.543$\pm$0.021*}} & \best{\textbf{0.454$\pm$0.020}} & \best{\textbf{0.081$\pm$0.037}} & \best{\textbf{0.674$\pm$0.015}} \\
\bottomrule
\end{tabular}
}

\end{table*}

\subsection{Generated-Example Augmentation at 200-Frame Horizon}
\label{Generated_Augmentation}
\begin{figure}[t]
    \centering
    \includegraphics[width=\linewidth]{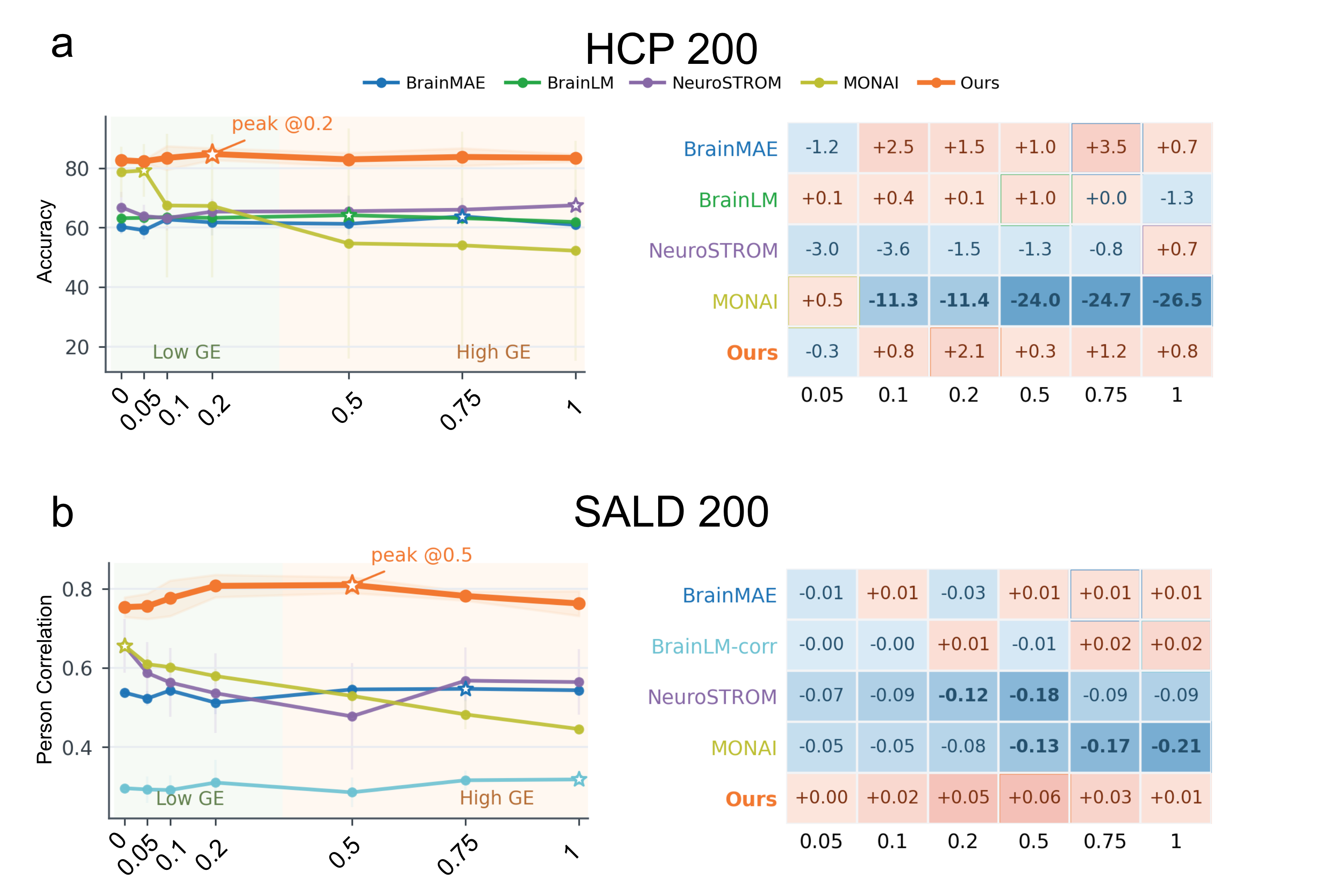}
    \caption{
    Additional generated-example (GE) augmentation results on HCP-200 and SALD-200.
    (a) HCP-200 gender classification accuracy under different GE ratios, with the corresponding performance changes relative to \(\mathrm{GE}=0\) shown on the right.
    (b) SALD-200 age regression performance measured by Pearson correlation under different GE ratios, with relative changes shown on the right.
    BrainWorld achieves consistently strong performance across GE ratios, whereas most baselines show limited gains or performance degradation when more generated samples are added.
    }
    \label{fig:supp_visualization_5}
\end{figure}

To further examine whether the augmentation effect observed in the main 400-frame setting also holds at a shorter generation horizon, we conduct an additional GE augmentation ablation on HCP-200 and SALD-200 in Fig.~\ref{fig:supp_visualization_5}. 
The real training set is kept fixed, and generated samples from the 200-frame setting are added at different GE ratios. 
For HCP-200, we evaluate gender classification accuracy, whereas for SALD-200, we evaluate age-regression performance using Pearson correlation. 
The line plots show the absolute downstream performance under each GE ratio, and the heatmaps summarize the corresponding performance changes relative to \(\mathrm{GE}=0\). 
Although the gains from 200-frame generated samples are generally weaker than those obtained with 400-frame generation in the main experiments, BrainWorld still provides consistent positive improvements and remains stronger than the competing baselines, while several baselines show limited gains or degradation as more generated samples are added.

\section{Details of Materials}

\subsection{Preprocessing}
\label{supp:processing}

\paragraph{fMRI preprocessing.}
All fMRI scans are converted to a common 4D voxel space before model training and evaluation. 
For datasets without released minimally preprocessed volumes, we perform standard preprocessing including motion correction, skull stripping, nuisance removal, and spatial normalization. 
All functional volumes are registered to the MNI152 template and resampled to 2~mm isotropic resolution. 
To reduce variability in acquisition protocols, BOLD time series are temporally interpolated to a unified TR of 0.72~s using cubic interpolation. 
Voxel intensities are normalized with background-aware Z-scoring, where normalization statistics are computed only over non-background brain voxels and background regions are kept fixed. 
Each 3D volume is then cropped or zero-padded to a fixed spatial size of \(96\times96\times96\). 
The resulting 4D sequence is divided into non-overlapping 40-frame windows, which are used as the basic input units for VAE encoding and diffusion-based future-window prediction. 
Due to the computational cost of large-scale fMRI preprocessing, for some datasets we use the subset of subjects that have completed preprocessing and quality control, rather than the entire public release. 
Within each dataset, the retained subjects are split at the subject level into training, validation, and test sets with a ratio of 7:1:2, unless otherwise specified by a task-specific evaluation protocol.

\paragraph{sMRI preprocessing.}
Structural MRI is processed separately to provide subject-level anatomical conditioning. 
For each T1-weighted scan, we use a unified T1 preprocessing pipeline consisting of N4 bias-field correction, skull stripping with HD-BET, and affine registration to the MNI152NLin2009cAsym 1-mm T1w template. 
The registered image is then reoriented to canonical space, non-finite values are set to zero, and the volume is resampled to the fixed 1~mm MNI reference grid with spatial size \(193\times229\times193\). 
The resulting single-channel T1 volume is encoded by the frozen structural MRI encoder to obtain a subject-level anatomical embedding, which serves as a stable structural prior for conditioning future fMRI generation.

\subsection{Datasets}
\label{supp:datasets}
\paragraph{Amsterdam Open MRI Collection (AOMIC).}
AOMIC contributes two young-adult cohorts to our pretraining pool. 
PIOP1~\cite{PIPO1_PIPO2} contains 168 participants (\(22.15 \pm 1.83\) years), with sex labels available for 161 participants (94 males and 67 females). 
PIOP2~\cite{PIPO1_PIPO2} contains 180 participants (\(21.99 \pm 1.83\) years), with sex labels available for 179 participants (98 males and 81 females). 
The discrepancy between the total number of participants and the reported sex counts is due to missing sex labels for a small number of participants. 
We treat the two releases as independent datasets because they were acquired as separate AOMIC studies.

\paragraph{Chinese Human Connectome Project (CHCP).}
CHCP~\cite{CHCP} is used to broaden the population coverage of the pretraining corpus beyond Western cohorts. We include 244 participants, with 117 males and 127 females, spanning 18--79 years of age. Its wider age range complements the young-adult HCP and AOMIC cohorts.

\paragraph{Imaging Chinese Young Brains (ISYB).}
ISYB~\cite{ISYB} provides an additional young-adult Chinese cohort. The subset used here includes 130 healthy Han Chinese participants aged 18--30 years (\(22.52\pm2.61\) years), with 98 females and 32 males. It is included as part of the multi-dataset pretraining corpus.

\paragraph{Adolescent Brain Cognitive Development (ABCD).}
ABCD~\cite{ABCD} introduces early-developmental samples into our dataset collection. We use 2400 participants from the 9--10-year-old baseline cohort. This dataset provides a pediatric contrast to the adult and lifespan datasets used elsewhere in the study.

\paragraph{Autism Brain Imaging Data Exchange (ABIDE).}
ABIDE~\cite{ABIDE} is used as an autism-related disease cohort for external evaluation. After our quality-control and filtering steps, the retained subset includes 609 participants from both autism spectrum disorder and typically developing control groups, with ages ranging from 6 to 58 years.

\paragraph{Human Connectome Project (HCP).}
HCP~\cite{HCP} provides the main high-quality young-adult reference cohort. We use 707 participants (\(28.79\pm3.53\) years) for pretraining, generation analysis, and downstream evaluation. HCP REST LR1, HCP-400, and HCP-200 in the main text denote different protocols or frame-length settings derived from this same cohort.

\paragraph{Parkinson's Progression Markers Initiative (PPMI).}
PPMI~\cite{PPMI} is used for Parkinson's disease-related evaluation. We retain 331 participants after quality control and label filtering, covering healthy controls, prodromal cases, and clinically diagnosed Parkinson's disease, with ages between 35 and 84 years.

\paragraph{Alzheimer's Disease Neuroimaging Initiative (ADNI).}
ADNI~\cite{ADNI} supports evaluation on Alzheimer's disease progression. The curated subset contains 497 participants with complete resting-state fMRI and diagnostic labels. The retained subjects span 55--90 years and include cognitively normal controls, mild cognitive impairment, and Alzheimer's disease groups.

\paragraph{Southwest University Adult Lifespan Dataset (SALD).}
SALD~\cite{SALD} is used for age-regression evaluation because of its broad adult age coverage. We include resting-state fMRI from 493 participants. SALD-400 and SALD-200 refer to different frame-length protocols used in the main experiments, not to separate datasets.

\paragraph{Brazilian High-Risk Cohort (BHRC).}
We use the preprocessed BHRC~\cite{BHRC} subset released through the Reproducible Brain Corpus (RBC) benchmark resource~\cite{rbc}, retaining 465 participants after preprocessing and label filtering. 
This dataset complements the HCP and SALD cohorts by providing an additional developmental evaluation setting.

\paragraph{Nathan Kline Institute (NKI).}
NKI~\cite{NKI} provides a lifespan neuroimaging cohort with resting-state fMRI and rich phenotypic annotations, making it suitable for evaluating individual-difference prediction across adulthood. 
We use the preprocessed NKI subset released through the Reproducible Brain Corpus (RBC) benchmark resource~\cite{rbc}. 
After preprocessing and label filtering, 717 participants are retained for age regression, and 331 participants with valid education labels are used for education-level classification. 
Education labels are grouped into three categories: primary education (Grades 1--6), secondary education (Grades 7--12), and higher
education (those with graduate-level degrees).

\paragraph{HCP Task.}
HCP Task~\cite{HCP} is treated separately from resting-state HCP because it provides task-evoked brain states. The benchmark covers seven task families and 23 experimental conditions, including working memory, motor, emotion, gambling, language, relational reasoning, and social cognition.

\paragraph{StudyForrest.}
StudyForrest~\cite{StudyForrest} is included as a naturalistic-stimulus dataset. The subset contains 20 participants aged 21--38 years (\(26.6\) years on average; 12 males, 8 females), with 7T fMRI acquired during a two-hour auditory movie stimulus. Each participant contributes 3,599 volumes.

\paragraph{CineBrain.}
CineBrain~\cite{gao_cinebrain_2025} is used as a naturalistic audiovisual dataset for evaluating stimulus-related brain dynamics. It contains simultaneous fMRI, EEG, and ECG recordings collected while participants watched narrative video clips from \textit{The Big Bang Theory}. The public description reports approximately six hours of audiovisual stimulation for each of six participants, making it suitable for long-form movie-driven fMRI generation and stimulus-conditioned analysis.

\paragraph{Emo-FilM.}
Emo-FilM~\cite{Emo_Film} provides a naturalistic emotion-elicitation setting based on short film viewing. The dataset contains fMRI and physiological recordings from 30 healthy participants watching 14 emotion-inducing films, together with dense emotion annotations from an independent group of raters.

\paragraph{HCP Movie.}
HCP Movie refers to the 7T movie-watching fMRI data from the HCP Young Adult project~\cite{HCP}. 
The movie runs were acquired after resting-state scans in the first and fourth 7T sessions, using the same 7T EPI sequence as resting-state fMRI. 
Participants watched short independent-film and Hollywood movie clips concatenated into runs of approximately 12--14 minutes. 
Due to the computational cost and preprocessing throughput of large-scale 7T movie fMRI, we use the subset that has completed preprocessing and quality control, consisting of 41 participants and 141 runs. 
We treat HCP Movie separately from resting-state HCP because it provides a controlled naturalistic-stimulus condition for evaluating cross-state and stimulus-driven long-horizon generation.

\paragraph{Healthy Brain Network (HBN).}
HBN~\cite{HBN} is included as a pediatric and adolescent neuroimaging cohort with rich phenotypic characterization. The project aims to build a large-scale biobank of participants aged 5--21 years, including multimodal brain imaging and behavioral, cognitive, and psychiatric assessments. In BrainWorld, HBN contributes developmental data outside the adult cohorts and supports evaluation of model robustness across younger populations.

\paragraph{1000 Functional Connectomes Project (FCON).}
The 1000 Functional Connectomes Project (FCON)~\cite{FCON} is a large multi-site resting-state fMRI resource. The original release aggregates over 1,200 resting-state fMRI datasets collected independently across 33 imaging sites. In BrainWorld, FCON is used to increase acquisition-site diversity and improve robustness to heterogeneous resting-state protocols.

\paragraph{Caltech Conte Center Dataset.}
The Caltech Conte Center~\cite{caltech} dataset provides multimodal neuroimaging and behavioral data from healthy adults. The release includes structural MRI, resting-state fMRI, movie-watching fMRI, and psychological assessments, with repeated acquisitions available for many participants. We use this dataset to add a behaviorally rich, multi-session adult cohort with both resting-state and naturalistic movie-viewing conditions.

\paragraph{Southwest University Longitudinal Imaging Multimodal Dataset (SLIM).}
The Southwest University Longitudinal Imaging Multimodal dataset (SLIM)~\cite{SLIM} is a longitudinal multimodal neuroimaging cohort from Southwest University. It includes structural MRI, resting-state fMRI, diffusion MRI, and behavioral measures collected across up to three time points over approximately three and a half years. We include SLIM to introduce longitudinal resting-state data and improve robustness to repeated-measure and test--retest variability.

\paragraph{Consortium for Reliability and Reproducibility (CoRR).}
The Consortium for Reliability and Reproducibility (CoRR)~\cite{CORR} is used as a test--retest-oriented resting-state resource. CoRR aggregates resting-state fMRI and diffusion imaging data from multiple laboratories to support reliability and reproducibility analyses in functional and structural connectomics.

\subsection{Condition Preparation for BrainWorld}
\label{supp:condition-preparation}

\paragraph{Overview of conditions.}
BrainWorld supports two types of conditions: intrinsic subject/context conditions and optional external stimulus conditions. 
Intrinsic conditions include a subject-level structural MRI embedding and a past-window functional connectivity embedding, whereas stimulus conditions include video and audio embeddings for naturalistic-stimulus datasets. 
These conditions are prepared by frozen modality-specific encoders and then projected by the condition encoder into token-level and global-level representations for the DiT backbone.

\paragraph{Intrinsic structural and dynamic-context conditions.}
BrainWorld uses two non-stimulus conditions to describe subject-specific anatomy and recent functional context. 
The first is an intrinsic structural condition derived from sMRI, which provides a stable anatomical prior for each subject. 
Specifically, each preprocessed T1-weighted MRI is encoded by a frozen BrainIAC~\cite{BrainIAC} encoder to obtain an MRI embedding. 
The second is a dynamic-context condition derived from the past fMRI window. 
We compute past functional connectivity (past FC) using the Schaefer-100~\cite{Schaefer} parcellation and extract an FC embedding with a frozen BrainMass~\cite{yang_brainmass_2024} encoder. 
The resulting MRI and past-FC embeddings are projected by the condition encoder into token-level and global-level representations for the DiT.

During training, we adopt classifier-free guidance (CFG)-style condition dropout~\cite{ho2022classifier}. 
When a condition is intentionally dropped, or when an MRI condition is unavailable for a subject, the corresponding condition is replaced by a learned null embedding. 
This design avoids requiring every dataset to provide all modalities, which is important for large-scale multi-dataset training where structural MRI, stimulus annotations, or other auxiliary conditions may be incomplete. 
It also enables BrainWorld to support conditional, partially conditional, and unconditional generation during inference.

\paragraph{Stimulus conditions.}
Video and audio are treated as optional external stimulus conditions for naturalistic-stimulus modeling. 
For the \(i\)-th fMRI window, we denote its BOLD interval as \([b_i,b_i+L)\). 
To compensate for the hemodynamic delay between external stimuli and BOLD responses, we use a fixed lag of \(\Delta=5\)~s and define the corresponding stimulus interval as
\[
I_i=
\left[
\max(0,b_i-\Delta),
\max(0,b_i+L-\Delta)
\right].
\]
Each lagged stimulus interval is divided into \(K=4\) consecutive clips, each with a nominal duration of 7.2~s. 
Video clips are encoded with V-JEPA 2.1 ViT-L/16~\cite{vjepa} by uniformly sampling 64 frames per clip, resizing/cropping frames to \(384\times384\), and mean-pooling the output tokens to obtain a clip-level video feature. 
Audio clips are resampled to 16~kHz, converted into 80-dimensional fbank features, encoded with W2v-BERT 2.0~\cite{wac}, and aggregated with masked mean pooling over valid frames to obtain a clip-level audio feature. 
Thus, each stimulus-conditioned fMRI window is paired with four video embeddings and four audio embeddings.

Naturalistic-stimulus datasets differ in how their stimulus timelines are defined. 
We therefore first construct a dataset-specific stimulus timeline and then apply the same 5-s lag and 7.2-s clip slicing rule. 
For CineBrain, multiple subjects watched common movie runs, so we build a shared run-level stimulus timeline and align each subject's fMRI segments to this common timeline. 
For StudyForrest~\cite{StudyForrest}, movie timing is reconstructed separately for each subject from the provided event logs; when a clip crosses a run boundary, the corresponding subsegments are extracted from adjacent runs and concatenated. 
For HCP Movie~\cite{HCP}, fMRI windows are generated with an overlapping-window protocol: each window contains 40 TRs, corresponding to 28.8~s, and adjacent windows are shifted by 17 TRs, corresponding to 12.24~s. 
Thus, neighboring fMRI windows may partially share stimulus content. 
Despite these dataset-specific timeline constructions, all naturalistic-stimulus datasets are converted into the same condition format: each fMRI window is paired with four lag-aligned video clips and four lag-aligned audio clips, each lasting 7.2~s.

Because stimulus datasets are much smaller than resting-state and disease datasets, we do not inject video/audio conditions during the initial mixed-state training stage. 
In this stage, naturalistic-stimulus datasets are trained together with resting-state and disease datasets, but their stimulus conditions are dropped and treated as absent. 
Emo-FilM is also trained in this way: we do not crop its corresponding stimulus clips, and instead treat it as a regular fMRI dataset without stimulus conditioning. 
After the initial mixed-state training, we further fine-tune BrainWorld on CineBrain and StudyForrest with video/audio conditions enabled, allowing the model to adapt to explicit stimulus-condition injection. 
HCP Movie is not used during this stimulus-condition fine-tuning stage and is reserved as an external naturalistic-stimulus evaluation dataset.

\paragraph{Condition injection.}
In the DiT backbone, MRI and past-FC embeddings provide intrinsic structural and dynamic-context conditions, while video and audio embeddings provide optional external stimulus conditions. 
MRI and FC embeddings are projected into token-level and global-level condition representations. 
Video and audio embeddings are encoded as four-token sequences, with an additional mean-pooled global vector for each modality. 
All available condition tokens are concatenated into a unified condition-token sequence and injected into each DiT block through cross-attention. 
In parallel, modality-level global vectors are fused by attention into a single global condition. 
This global condition is combined with the diffusion timestep embedding to generate adaptive shift, scale, and gate parameters for each DiT block, modulating the self-attention, cross-attention, and MLP branches. 
Therefore, BrainWorld uses a three-level conditioning mechanism: token-level cross-attention, global-level modality fusion, and block-level adaptive modulation. 
Missing or dropped modalities are represented by learned null tokens and null global embeddings, enabling robust training and inference under incomplete conditions.

\subsection{Baseline Models}
\label{supp:baseline}
We compare BrainWorld with representative fMRI foundation models. These baselines are used for two evaluation purposes: 4D fMRI generation and downstream task prediction. BrainLM and NeuroSTORM are evaluated in both settings, serving as baselines for generation quality and downstream transfer performance, whereas MONAI is used only as a generation baseline. For each baseline, we follow the released implementation or reported architecture when available, and adapt the prediction head to the corresponding task setting used in our experiments. Unless otherwise specified, all baselines are evaluated under the same train/validation/test split as BrainWorld.

\paragraph{BrainLM.}
BrainLM~\cite{caro_brainlm_2023} is an ROI-level fMRI foundation model based on masked autoencoding. It represents each fMRI recording as AAL-424 ROI time series and learns to reconstruct masked brain activity tokens using a Transformer encoder--decoder. BrainLM was pretrained on 77,298 fMRI recordings from UK Biobank and HCP, corresponding to approximately 6,700 hours of fMRI data. For downstream fine-tuning, we use a batch size of 64, a learning rate of \(1\times10^{-5}\), and 30 training epochs.

\paragraph{Brain-JEPA.}
Brain-JEPA~\cite{dong2024brain} follows a joint-embedding predictive learning paradigm for fMRI dynamics. It uses 450 ROIs obtained from Schaefer-400 cortical parcels and Tian Scale-III subcortical regions. Instead of reconstructing masked signals directly, Brain-JEPA predicts target latent embeddings from visible context under spatial and temporal masking. The model is pretrained on UK Biobank data and then adapted to downstream tasks. In our setup, Brain-JEPA is trained or fine-tuned with a batch size of 16, a learning rate of \(4\times10^{-4}\), and 50 epochs.

\paragraph{BrainMass.}
BrainMass~\cite{yang_brainmass_2024} is a functional-connectivity foundation model built on Schaefer-100 brain networks. It learns network-level representations through self-supervised objectives that combine masked ROI modeling with latent representation alignment. The released model was pretrained on 26 datasets with 64,584 subjects, including UK Biobank, HCP, and ADHD-200. In our downstream experiments, BrainMass is fine-tuned with a batch size of 64, a learning rate of \(2\times10^{-4}\), and 100 training epochs.

\paragraph{Brain-DiT.}
Brain-DiT~\cite{xia2026brain} is included as a diffusion-based ROI-level foundation model baseline for downstream representation learning. It was pretrained on 349,898 fMRI sessions from 24 datasets, covering multiple brain states including resting, task-evoked, naturalistic-stimulus, disease, and sleep states. The main Brain-DiT model uses Schaefer-1000 ROI time series as input, with each fMRI window represented as an ROI--time matrix. An additional AAL424 variant is used in the original study to align with the atlas setting of BrainLM. During pretraining, Brain-DiT adds noise to an ROI--time window at a sampled diffusion timestep and trains a Diffusion Transformer to predict the \(v\)-target under a 1000-step DDPM schedule. Demographic and clinical metadata, including age, sex, and diagnosis, are embedded as conditioning variables and injected into DiT blocks through AdaLN-Zero, with classifier-free condition dropout used during training.

\paragraph{SLIMBrain.}
SLIMBrain~\cite{wang_slim-brain_2026} is used as an efficient atlas-free voxel-level foundation model baseline. It first uses a lightweight temporal extractor to score and select informative fMRI windows, and then feeds only the selected windows into a 4D Hiera-JEPA encoder for fine-grained voxel-level representation learning. This selective-compute design reduces memory and data requirements while preserving voxel-level spatial details. We include SLIMBrain as an atlas-free representation baseline complementary to ROI-level models.

\paragraph{SwiFT.}
SwiFT~\cite{kim_SWIFT_2023} is a voxel-level 4D fMRI Transformer based on shifted-window attention. Unlike ROI-based methods, it takes raw 4D fMRI volumes as input and performs local spatiotemporal attention within 4D windows, making it a strong end-to-end baseline for volume-based fMRI representation learning. In our experiments, SwiFT is fine-tuned using a batch size of 16, a learning rate of \(1\times10^{-6}\), weight decay of 0.01, and 30 training epochs.

\paragraph{NeuroSTORM.}
NeuroSTORM~\cite{wang_towards_2025} is a voxel-level fMRI foundation model designed for direct learning from raw 4D fMRI volumes. Its backbone combines shifted-window processing with Mamba-style sequence modeling to reduce the cost of long-range spatiotemporal modeling. The model is pretrained on large-scale fMRI data from more than 50,000 subjects across UK Biobank, ABCD, and HCP. It also uses a spatiotemporal redundancy dropout strategy to prevent masked pretraining from relying on local redundancy. We include NeuroSTORM as a recent large-scale volume-based foundation model baseline.

\paragraph{Brain Harmony.}
Brain Harmony~\cite{harmony} is included as a multimodal structure--function foundation model baseline. It separately encodes T1-weighted structural MRI and fMRI dynamics, and then fuses the modality-specific latents into compact 1D brain-hub tokens. For the functional branch, Brain Harmony uses Schaefer-400 ROI time series and introduces temporal adaptive patch embedding to handle heterogeneous TRs. For the structural branch, T1-weighted volumes are encoded with a 3D masked autoencoding framework. During multimodal fusion, the structural and functional latents are compressed into shared hub tokens, which are then used for downstream prediction. Brain Harmony was pretrained on UK Biobank and ABCD, using 64,594 T1-weighted volumes and 70,933 fMRI time series. In our experiments, we use Brain Harmony as the main representation-level F+S fusion baseline to compare against BrainWorld's generative structure-function fusion. Brain Harmony fine-tunes downstream tasks for 50 epochs using AdamW with a learning rate of 5e-4, weight decay of 0.05, layer-wise learning-rate decay of 0.65, 5 warmup epochs, and cosine warmup-decay scheduling. Its linear-probe setting uses AdamW for 100 epochs with a learning rate of 1e-3, typically with batch size 8 in the public example scripts.

\paragraph{Omni-fMRI.}
Omni-fMRI~\cite{wang2026omni} is used as an atlas-free voxel-level foundation model baseline. It avoids ROI parcellation and directly models 4D fMRI volumes with a dynamic patching strategy, where background regions are pruned and foreground regions are assigned adaptive patch sizes according to local spatiotemporal complexity. This design reduces token length while preserving fine-grained voxel information. The model was pretrained on 49,497 fMRI sessions across nine datasets and benchmarked on 11 datasets. For downstream adaptation, we fine-tune Omni-fMRI for 30 epochs using AdamW with a base learning rate of \(1\times10^{-5}\), a head learning rate of \(1\times10^{-4}\), weight decay of 0.05, layer decay of 0.75, 2 warmup epochs, linear warmup-decay scheduling, and a total batch size of 32.

\paragraph{BrainIAC.}
BrainIAC~\cite{BrainIAC} is included as a structural MRI foundation model baseline. Unlike fMRI-specific baselines that operate on BOLD dynamics or ROI time series, BrainIAC is pretrained on 3D brain MRI volumes and learns general-purpose anatomical representations through self-supervised contrastive learning. It was pretrained on 32,015 multiparametric brain MRI scans and evaluated within a larger collection of 48,965 brain MRIs spanning diverse neurological conditions and MRI sequences. The model uses a SimCLR-style objective, where augmented 3D MRI patches from the same anatomical region form positive pairs and patches from other samples serve as negatives. Based on backbone benchmarking, the original study adopted a ViT-B encoder as the final BrainIAC backbone.

In our experiments, BrainIAC is used to represent the structural MRI modality. Preprocessed T1-weighted MRI volumes are passed through the frozen BrainIAC encoder to extract subject-level anatomical embeddings, which serve either as structural-only features or as MRI conditions in BrainWorld. We include BrainIAC to provide a strong structural representation baseline and to support comparison against multimodal methods that combine structural and functional information. BrainIAC fine-tunes downstream tasks for 30 epochs using AdamW with a single learning rate of 8e-4, weight decay of 1e-4, batch size 8, and a CosineAnnealingWarmRestarts scheduler under mixed-precision GPU training. Unlike Brain-Harmony, the current BrainIAC downstream code does not expose separate backbone/head learning rates, layer-wise decay, or warmup scheduling.

\paragraph{MONAI-based latent diffusion baseline.}
\label{supp:monai_intro}
We use a MONAI-based latent diffusion model as a generation baseline for 4D fMRI future-window prediction~\cite{pinaya2023generative}. 
Specifically, we use MONAI GenerativeModels to implement the Stage-1 VQ-VAE tokenizer, but do not use any off-the-shelf MONAI pretrained weights; all components are trained on fMRI data. 
This baseline provides a medical-image generative comparison under the same next-window autoregressive setting as BrainWorld. 
Unlike BrainWorld, whose VAE is designed for 4D fMRI latent modeling, the MONAI tokenizer is a frame-wise 3D VQ-VAE.

In Stage 1, the VQ-VAE is trained on individual 3D fMRI frames. 
Each 4D clip is standardized to \(96\times96\times96\times40\), reordered from \texttt{xyzt} to \texttt{txyz}, and normalized with nonzero-voxel z-scoring. 
The trained VQ-VAE compresses each 3D frame into a latent grid of size \(4\times24\times24\times24\). 
For a 40-frame clip, the resulting frame-wise latents form a clip-level latent tensor of shape \([40,4,24,24,24]\). 
For the Stage-2 3D DiT, we merge the temporal and latent-channel dimensions, yielding
\[
x_0 \in \mathbb{R}^{160\times24\times24\times24}.
\]

In Stage 2, MONAI uses a conditional 3D DiT to generate the next-window latent. 
For each target window, the immediately preceding latent window from the same subject, session, and sequence is used as the anchor condition. 
The noisy target latent is divided into non-overlapping \(4\times4\times4\) patches, producing \(216\) spatial tokens. 
The DiT uses width 768, depth 12, 12 attention heads, and an MLP ratio of 4.0. 
To follow the same autoregressive prediction protocol as BrainWorld, the model is conditioned on the anchor latent, an available dataset/state label, and a past-window FC embedding computed from the anchor fMRI window. 
The anchor latent and FC embedding are encoded into both global vectors for AdaLN-Zero modulation and condition tokens for cross-attention, allowing temporal context and FC-based functional information to influence the denoising process.

The diffusion objective follows \(v\)-prediction with a 1000-step DDPM schedule and a scaled-linear beta schedule. 
We train MONAI with AdamW using learning rate \(1\times10^{-4}\), weight decay \(0.01\), and EMA decay \(0.9999\). 
Condition-label dropout is applied for robustness, and samples without FC features are handled using an explicit \texttt{has\_fc} indicator and a learned null FC condition. 
During inference, MONAI starts from Gaussian noise, generates the target latent conditioned on the anchor latent, available label condition, and FC embedding, and then decodes the generated latent back to 4D fMRI space using the trained VQ-VAE decoder.

\subsection{Configurations}

This subsection summarizes the configuration details required to reproduce BrainWorld and the comparison baselines, including architectural parameters, condition encoders, training schedules, rollout settings, and downstream fine-tuning and linear-probing protocols. Tables~\ref{tab:ddit-configs} and~\ref{tab:wfvae-configs} summarize the main hyperparameter settings used in BrainWorld, including the conditional latent DiT backbone, the default downstream probing protocol, and the VAE variants.

\begin{table}[t]
\centering
\small
\caption{Key hyperparameter settings of the Conditional Latent DiT and its full-finetuning protocol. BS denotes batch size.}
\label{tab:ddit-configs}
\renewcommand{\arraystretch}{1.08}
\begin{tabular}{p{0.44\linewidth} p{0.46\linewidth}}
\hline
\textit{Config} & \textit{Value} \\
\hline
\multicolumn{2}{c}{\textbf{Conditional Latent DiT Pretraining}} \\
\hline
optimizer & AdamW \\
learning rate / weight decay & $1\times10^{-4}$ / 0.02 \\
training epochs & 50 \\
total batch size & $8$ GPUs $\times$ $4$ BS $=32$ \\
diffusion steps / schedule & 1000 / cosine \\
prediction target & $v$-prediction, with \texttt{x0\_aux\_weight} $=0.05$ \\
latent input shape & $[40, 16, 10, 12, 10]$ \\
spatial patch size & $[2, 4, 2]$ \\
tokens per window & $40 \times 5 \times 3 \times 5 = 3000$ \\
transformer depth / heads / dim & 16 / 8 / 512 \\
total parameters & 116,516,160 \\
\hline
\multicolumn{2}{c}{\textbf{Full-finetuning Protocol}} \\
\hline
optimizer & AdamW \\
DiT / aggregator / head learning rates & $1\times10^{-5}$ / $1\times10^{-4}$ / $1\times10^{-4}$ \\
DiT / aggregator / head weight decay & $1\times10^{-4}$ / 0 / 0 \\
training epochs / patience & 40 / 10 \\
total batch size & $5$ GPUs $\times$ $32$ BS $=160$ \\
gradient clipping & global norm 1.0 \\
embedding timesteps & $[0, 50, 100, 150]$ \\
captured transformer layers & $[-6, -4, -2, -1]$ layers \\
aggregator heads / dropout & 4 / 0.1 \\
\hline
\end{tabular}
\end{table}

\begin{table}[t]
\centering
\small
\caption{Hyperparameter settings of the BrainWorld VAE configuration. BS denotes batch size.}
\label{tab:wfvae-configs}
\renewcommand{\arraystretch}{1.08}
\begin{tabular}{p{0.44\linewidth} p{0.46\linewidth}}
\hline
\textit{Config} & \textit{Value} \\
\hline
\multicolumn{2}{c}{\textbf{VAE}} \\
\hline
input shape convention & $[B, 1, T, D, H, W]$ \\
canonical crop size & spatial crop $[80, 96, 80]$, input $T=40$ \\
stage channels & $[16, 32, 64, 96]$ \\
latent dimension & 16 \\
residual blocks per stage & 2 \\
spatial downsampling stages & 3 \\
spatial-only frame chunk size & 8 \\
spatial kernel size & 5 \\
temporal kernel size & 5, defined but unused in the bypassed path \\
dropout & 0.0 \\
temporal residual scale & 0.1 \\
wavelet energy channels & 8 \\
wavelet fusion stages & \texttt{enc2, dec2} \\
log-variance clamp & $[-6.0, 4.0]$ \\
resulting latent shape & $[B, 40, 16, 10, 12, 10]$ \\
optimizer & AdamW \\
optimizer momentum & $\beta_1,\beta_2 = 0.9, 0.999$ \\
learning rate schedule & constant learning rate, no explicit scheduler \\
learning rate & $3 \times 10^{-4}$ \\
weight decay & $1 \times 10^{-3}$ \\
gradient clipping & global norm 1.0 \\
mixed precision & \texttt{bf16} AMP \\
per-GPU batch size & 2 \\
total batch size & $6$ GPUs $\times$ $2$ BS $=12$ \\
training epochs & 5 \\
reconstruction loss weights & foreground $=1.0$, background $=0.05$, bg-zero $=0.02$ \\
auxiliary loss weights & wavelet $=0.05$, temporal $=0.0$, $\beta_{\mathrm{KL}}=1\times10^{-4}$ \\
early stopping & enabled, patience 8, min delta $=1\times10^{-4}$ \\
total parameters & 3,557,829 \\
\hline
\end{tabular}
\end{table}

\subsection{Computational Resources}
\label{supp:compute}

All experiments were conducted on Linux GPU servers using PyTorch with mixed-precision training. 
BrainWorld VAE pretraining and conditional latent DiT training were performed on NVIDIA L40 GPUs with 48~GB memory using distributed data parallelism when applicable. 
The Stage-I VAE contains 3.56M parameters and was trained with a per-GPU batch size of 2 on 6 L40 GPUs, giving a total batch size of 12. 
The Stage-II conditional latent DiT contains 116.5M parameters and was trained with a total batch size of 32 on 8 L40 GPUs using distributed data parallelism, as summarized in Table~\ref{tab:ddit-configs}. 
Stage-II DiT pretraining took approximately 2 days, corresponding to roughly 384 GPU-hours. 
Unless otherwise specified, mixed precision was enabled to reduce memory usage and improve training efficiency.

Downstream full fine-tuning was conducted using the protocol in Table~\ref{tab:ddit-configs}, with a total batch size of 160 across 5 GPUs. 
Linear probing and baseline fine-tuning were substantially cheaper and were run with smaller GPU allocations depending on the memory requirement of each model.

For generation evaluation, future 4D fMRI windows were sampled autoregressively in latent space and decoded by the frozen VAE decoder. 
A 400-frame trajectory requires 10 sequential 40-frame rollout steps. 
For the full 400-frame evaluation pipeline, including autoregressive latent prediction, decoding, and ROI-level metric computation, the end-to-end runtime was approximately 95.5 hours.

\subsection{Evaluation Details}
\label{supp:evaluation-details}

\paragraph{Generative forecasting.}
For generative forecasting, we compare BrainWorld with two categories of baselines. 
The first category includes fMRI foundation models, including BrainMAE, BrainLM, and NeuroSTORM. 
Since these models are not originally designed for long-horizon generation, we adapt them with an autoregressive mask-then-fill protocol, where predicted frames are iteratively reused as context for subsequent prediction steps. 
We further finetune BrainMAE, BrainLM, and NeuroSTORM on HCP and CineBrain under the same forecasting protocol to ensure fair comparison. 
The second category is a MONAI-based latent diffusion baseline implemented with a conditional 3D DiT backbone. 
Following Seo et al.~\cite{seo_scalable_2025}, we stack frame-wise latent features along the channel dimension to adapt the 3D latent diffusion framework to 4D fMRI future-window prediction.

For fair comparison with ROI-level forecasting baselines, BrainWorld's voxel-level predictions are parcellated into ROI time series using the corresponding atlas before metric computation. 
MSE is computed over the ROI-by-time matrices, and Pearson correlation is computed as the average temporal correlation across ROIs. 
For voxel-level visualization, predictions are decoded back to 4D fMRI volumes using the frozen VAE decoder and compared with ground-truth volumes at matched horizons.

\paragraph{Generated-example augmentation.}
For generated-example (GE) augmentation, the real training set is kept fixed and generated samples are added according to the GE ratio. 
\(\mathrm{GE}=0\) denotes training only with real samples, whereas \(\mathrm{GE}=1.0\) adds the same number of generated samples as real training samples. 
Generated examples inherit subject-level labels and sMRI conditions from their conditioning real samples, and validation and test sets contain only real samples. 
To avoid confounding augmentation with additional optimization, all GE ratios use the same number of optimization steps and the same validation-based early stopping protocol.

\paragraph{Downstream transfer.}
For downstream transfer, we evaluate both full fine-tuning and frozen-backbone linear probing. 
Full fine-tuning updates the DiT backbone, feature aggregator, and downstream head, whereas linear probing keeps the backbone frozen and trains only the lightweight downstream readout. 
We compare functional-only (F), structural-only (S), and multimodal (F+S) baselines. 
To isolate the effect of the fusion mechanism, we additionally include BrainWorld$_{\mathrm{F+S\text{-}concat}}$, a same-backbone representation-level fusion baseline that concatenates fMRI features and independently extracted sMRI embeddings before the downstream head. 
BrainWorld$_{\mathrm{F+S\text{-}prior}}$ instead injects sMRI through the DiT structural-prior pathway.

For classification tasks, we report accuracy and F1 score. 
For regression tasks, we report MSE and Pearson correlation. 
Age labels are \(z\)-score normalized using statistics from the training split, and MSE is reported in the normalized label space.

\paragraph{Statistical reporting.}
All downstream results are reported as mean \(\pm\) standard deviation over three runs. 
Given the limited number of runs, we report Cohen's \(d\) as an effect-size measure rather than relying on significance testing. 
Asterisks in the main tables denote a large favorable effect size with Cohen's \(d\geq0.8\), computed in the favorable direction against the best external baseline for the same task and metric.